\documentclass[runningheads,a4paper,10pt]{llncs}

\usepackage{times}
\usepackage{amsmath}
\usepackage{amssymb}
\usepackage{url}  %Required
\usepackage{graphicx}  %Required

\usepackage{caption}

\usepackage{multirow}
\usepackage{algorithm}
\usepackage{algpseudocode}
\usepackage{enumitem}

\usepackage{tabularx}
\DeclareGraphicsExtensions{.png}
\usepackage[caption=false]{subfig}

\usepackage{tikz}
\usetikzlibrary{shapes,calc,fit,matrix,chains,positioning,decorations.pathreplacing,arrows}

\DeclareMathOperator*{\argmin}{argmin}

\newcommand{\keywords}[1]{\par\addvspace\baselineskip
\noindent\keywordname\enspace\ignorespaces#1}

\begin{document}

\mainmatter  % start of an individual contribution

\title{Cost-Sensitive Reference Pair Encoding for Multi-Label Learning}

%\titlerunning{Lecture Notes in Computer Science: Authors' Instructions}

\author{Yao-Yuan Yang
\and Kuan-Hao Huang \and Chih-Wei Chang \and Hsuan-Tien Lin}
\institute{CSIE Department, National Taiwan University \\
\{b01902066,r03922062\}@ntu.edu.tw, cwchang@cs.cmu.edu, htlin@csie.ntu.edu.tw
}
%\institute{Springer-Verlag, Computer Science Editorial,\\
%Tiergartenstr. 17, 69121 Heidelberg, Germany\\
%\url{http://www.springer.com/lncs}}


%\toctitle{Lecture Notes in Computer Science}
%\tocauthor{Authors' Instructions}
\maketitle

\begin{abstract}

Label space expansion for multi-label classification (MLC) is a methodology that encodes the original label vectors to higher dimensional codes before training and decodes the predicted codes back to the label vectors during testing. The methodology has been demonstrated to improve the performance of MLC algorithms when coupled with off-the-shelf error-correcting codes for encoding and decoding. Nevertheless, such a coding scheme can be complicated to implement, and cannot easily satisfy a common application need of cost-sensitive MLC---adapting to different evaluation criteria of interest. In this work, we show that a simpler coding scheme based on the concept of a reference pair of label vectors achieves cost-sensitivity more naturally. In particular, our proposed cost-sensitive reference pair encoding (CSRPE) algorithm contains cluster-based encoding, weight-based training and voting-based decoding steps, all utilizing the cost information. Furthermore, we leverage the cost information embedded in the code space of CSRPE to propose a novel active learning algorithm for cost-sensitive MLC. Extensive experimental results verify that CSRPE performs better than state-of-the-art algorithms across different MLC criteria. The results also demonstrate that the CSRPE-backed active learning algorithm is superior to existing algorithms for active MLC, and further justify the usefulness of CSRPE.  
\keywords{Multi-label Classification, Cost-sensitive, Active Learning}
\end{abstract}

\section{Introduction} \label{introduction}
The \textit{multi-label classification} (MLC) problem aims to map an instance to
multiple relevant labels~\cite{katakis2008multilabel,liu2004active}, which
matches the needs of many real-world applications, such as object detection and news classification.
%\cite{katakis2008multilabel,elisseeff2001kernel}.
%\cite{katakis2008multilabel,DBLP:conf/ismir/TrohidisTKV08,elisseeff2001kernel}.
Different applications generally require evaluating the performance of
MLC algorithms with different criteria, such as the
Hamming loss, 0/1 loss, Rank loss, and F1 score
\cite{DBLP:reference/dmkdh/TsoumakasKV10}.

Most existing MLC algorithms are designed to optimize one or few criteria.
For instance, \textit{binary relevance} (BR)
\cite{DBLP:reference/dmkdh/TsoumakasKV10} learns a binary classifier per
label to predict its relevance, and naturally optimizes the Hamming loss.
\textit{Classifier chain} (CC) \cite{read2011classifier} extends BR by
ordering the labels as a chain and using earlier labels of the chain to improve the per-label prediction,
and optimizes the Hamming loss like BR.
\textit{Label powerset} (LP) \cite{DBLP:reference/dmkdh/TsoumakasKV10} optimizes the 0/1 loss by solving
a multi-class classification problem that treats
each label combination as a hyper-class.
These \textit{cost-insensitive} algorithms cannot easily adapt to different criteria,
and may suffer from bad performance when evaluated with other criteria.

\textit{Cost-sensitive MLC} (CSMLC) algorithms are able to adapt to different criteria
more easily. In particular, CSMLC algorithms take the criterion as an additional piece of input data and aim to optimize
the criterion during the learning process.
Two state-of-the-art CSMLC algorithms are
\textit{probabilistic classifier chain} (PCC)
\cite{DBLP:conf/icml/DembczynskiCH10} and
\textit{condensed filter tree} (CFT) \cite{DBLP:conf/icml/LiL14}.
PCC estimates the conditional probability of the labels to infer the Bayes-optimal
decision with respect to the given criterion.
While PCC can tackle any criterion in principle, the Bayes-optimal inference step
can be time-consuming unless an efficient inference rule of the criterion is derived
in advance.
CFT can be viewed as an extension of CC for CSMLC by re-weighting each example with respect to the criterion
when training each binary classifier. Nevertheless, the re-weighting step depends on going back and forth within the chain,
making CFT still somewhat time-consuming and hardly parallelizable.

The \textit{multi-label error-correcting code} (ML-ECC) \cite{CF2013}
framework is a more sophisticated algorithm that goes beyond the per-label
classifiers to improve classification performance.
ML-ECC uses error-correcting code (ECC) to transform the original MLC problem into
a bigger MLC problem by adding error-correcting labels during encoding.
Classifiers on those labels, much like ECC for communication, can be used
to correct prediction errors made from the original per-label classifiers and
improve MLC performance.
While ML-ECC is successful in terms of the Hamming loss and 0/1 loss
\cite{CF2013}, it is not cost-sensitive and cannot easily adapt to other evaluation criteria.
In fact, extending ML-ECC for CSMLC problem appears to be highly non-trivial
and has not yet been deeply studied.

In this work, we study the potential of ECC for CSMLC by considering a special
type of ECC, the one-versus-one (OVO) code, which is a popular code for multi-class classification~\cite{DBLP:conf/acml/Lin14}.
We extend the OVO code to a cost-sensitive code,
\textit{cost-sensitive reference pair encoding} (CSRPE), which preserves
the information of the criterion in each code-bit during encoding.
We further propose a method to convert the criterion into instance weights during training,
and a method to take the criterion into account during decoding.
To make the whole CSRPE algorithm efficient enough
to deal with exponentially many possible label vectors, we study the possibility of sampling the code-bits
and zooming into a smaller subset of label vectors during prediction.
The resulting algorithm is as efficient as a typical random forest (when coupled with decision trees) in
training, and can be easily implemented in parallel.
Extensive experimental results demonstrate that CSRPE
outperforms existing ML-ECC
algorithms and the state-of-the-art CSMLC algorithms across different
criteria.

In addition, based on the proposed CSRPE, we design a novel algorithm for
\textit{multi-label active learning} (MLAL).
Retrieving ground-truth labels is usually expensive in real-world applications
\cite{liu2004active}.
The goal of MLAL is to actively query the labels for a small
number of instances while maintaining good test MLC performance.
Nevertheless, current MLAL algorithms
\cite{brinker2006active,yang2009effective,li2013active} are not capable of
taking the evaluation criterion into consideration when querying.
In this paper, we formulate the \textit{cost-sensitive multi-label active
learning} (CSMLAL) setting, and propose a novel algorithm that leverages the code
space computed by CSRPE to conduct cost-sensitive querying.
Experimental results justify that the proposed algorithm is superior to other
state-of-the-art MLAL algorithms.

This paper is organized as follows.
First, we define CSMLC problem formally and introduce the ML-ECC framework
in Section \ref{preliminary}.
Our proposed CSRPE algorithm is described in Section \ref{proposed}.
In Section \ref{csmlal}, we define the CSMLAL problem and solve it with a
novel algorithm based on CSRPE.
The empirical studies of both CSRPE and its active learning extension are
presented in Section \ref{experiments}
\footnote{Code for multi-label classification and active learning
will be available at
\url{https://github.com/yangarbiter/multilabel-learn}
and \url{https://github.com/ntucllab/libact}, respectively}.
Finally, we conclude the paper in Section \ref{conclusion}.

\section{Preliminary}\label{preliminary}

The goal of a MLC problem is to map the
feature vector $\mathbf{x} \in \mathcal{X} \subseteq \mathbb{R}^d$
to a label vector $\mathbf{y} \in \mathcal{Y} \subseteq \{0,
1\}^K$, where $\mathbf{y}[k] = 1$ if and only if the $k$-th bit is relevant.
During training, MLC algorithms use the training dataset $\mathcal{D} = \{(\mathbf{x}^{(n)},
\mathbf{y}^{(n)})\}_{n=1}^N$
to learn a classifier $f\colon \mathcal{X}~\to~\mathcal{Y}$.
During testing, for any test example $(\mathbf{x}, \mathbf{y})$ drawn from
the distribution that generated $(\mathbf{x}^{(n)}, \mathbf{y}^{(n)})$, the prediction
$f(\mathbf{x})$ is evaluated with a cost
function $C\colon \mathcal{Y} \times \mathcal{Y} \to \mathbb{R}$, where
$C(\mathbf{y}, \mathbf{\hat{y}})$ represents the penalty of predicting $\mathbf{y}$ as $\mathbf{\hat{y}}$.
The objective of MLC algorithms is to minimize the expected cost
$\mathbb{E}_{(\mathbf{x}, \mathbf{y})}[C(\mathbf{y}, f(\mathbf{x}))]$.

Traditional MLC algorithms are designed to optimize one or few cost functions.
These algorithms may suffer from bad performance when other cost functions
are used.
On the contrary, \textit{cost-sensitive multi-label classification} (CSMLC)
algorithms take the cost function as an additional input and learn a
classifier $f$ from both $\mathcal{D}$ and $C$.
Classifier $f$ should adapt to different $C$ easily.

%In this work, we focus on the \textit{multi-label error-correcting code}
%(ML-ECC) \cite{CF2013} framework and its possible extension to CSMLC.
The \textit{multi-label error-correcting code} (ML-ECC) \cite{CF2013}
framework is originally designed to optimize one cost function (the 0/1
loss).
ML-ECC borrows the error-correcting code (ECC) from the communication domain.
ML-ECC views the label vectors $\mathbf{y}^{(n)}$ as bit strings and encodes
them to longer codes $\mathbf{b}^{(n)} = enc(\mathbf{y}^{(n)})$ with some ECC
encoder $enc\colon \mathcal{Y} \to \{0, 1\}^M$, where $M$ is the code
length.
An MLC classifier $h$ is trained on $\{(\mathbf{x}^{(n)}, \mathbf{b}^{(n)})\}$
to predict the codes instead of the label vectors.
The code-bits store redundant information about the label vector to recover
the intended label vector even when some bits are mispredicted by $h$.
In prediction, the corresponding ECC decoder $dec\colon \{0, 1\}^M \to \mathcal{Y}$,
is used to convert the predicted vector from $h$ back to the label vector
$f(\mathbf{x}) = dec(h(\mathbf{x}))$.
In other words, ML-ECC learns the classifier $f = dec \; \circ \; h$.
Such an ECC decoder is often designed based on special nearest-neighbor search steps in the
code space~\cite{CF2013}.
%That is, the label vector whose code is closest to the predicted vector from
%$h$ is taken as the final prediction.

In the original work of ML-ECC \cite{CF2013}, several encoder/decoder choices
are discussed and experimentally evaluated.
Nevertheless, none of them take the cost information into account.
In fact, to the best of our knowledge, there is currently no work that deeply studies the potential of ECC for CSMLC. Next, we illustrate
our ideas on making a special ECC cost-sensitive.
%In this work, we study the design of the cost-sensitive ECC.
%We want to design the encoding function and decoding function with respect to
%the cost function, and preserve the information of the cost function in the
%encoded vectors.

\section{Proposed Approach} \label{proposed}

We start from a special cost-insensitive ECC, the \textit{one-versus-one}
(OVO) code. The OVO code is the core of the OVO meta-algorithm for
\textit{multi-class
  classification}~(MCC). The meta-algorithm trains many binary classifiers, each
representing the duel between \textit{two} of the classes, and let the binary classifiers vote for the majority decision for MCC.

To study the OVO code for MLC, we can na{\"i}vely follow the label powerset algorithm \cite{DBLP:reference/dmkdh/TsoumakasKV10} to reduce the MLC problem to MCC and then apply the OVO meta-algorithm to further reduce MCC to binary classification. As a consequence,
each label vector $\mathbf{y} \in \mathcal{Y}$ is simply treated as a distinct hyper-class,
and each binary classifier within the OVO meta-algorithm represents a duel between \textit{two} label vectors.
More specifically, the $i$-th classifier is associated with two label vectors~$\mathbf{y}^{i}_{\alpha}$ and $\mathbf{y}^{i}_{\beta}$, called the reference label vectors.
There are $\binom{2^{K}}{2}$ such classifiers, each can be trained with examples in $D$ that match either
$\mathbf{y}^{i}_{\alpha}$ and $\mathbf{y}^{i}_{\beta}$. During prediction,
the $\binom{2^{K}}{2}$ binary classifiers can then vote for all the label
vectors $\in \mathcal{Y}$ towards the majority decision.

The steps of applying OVO to MLC above can be alternatively described as a special ML-ECC algorithm,
similar to how OVO is viewed as a special ECC for MCC~\cite{allwein2001reducing}.
OVO as ML-ECC encodes each label vector to a code of length $\binom{2^K}{2}$ with the following encoder
%\begin{equation}% \label{eq:ovo_enc}
$enc_{ovo}(\mathbf{y})[i] =
  \begin{cases}
    1       & \quad \text{if } \mathbf{y} = \mathbf{y}^{i}_{\alpha}\\
    0       & \quad \text{if } \mathbf{y} = \mathbf{y}^{i}_{\beta}\\
    0.5      & \quad \text{otherwise}
  \end{cases}
$.
%\end{equation}
%The $i$-th bit in the code represents whether the label vector matches either
%of the reference vectors, with the special ``bit'' value of $0.5$ for
%representing other irrelevant vectors.
The $i$-th bit in the code represents whether the label vector matches either
of the reference vectors.
The special ``bit'' value of $0.5$ represents other irrelevant
label vectors.
Then, decoding based on majority voting is equivalent to
nearest-neighbor search in the code space over all possible encoded $\mathbf{y} \in \mathcal{Y}$ in terms of the
Hamming distance ($d_{ham}$), as the Hamming distance is a linear function of the vote that each $\mathbf{y}$ gets.
More precisely, denote the predicted code as $\mathbf{\hat{b}} = h(\mathbf{x})$,
the decoder of OVO is simply
$dec_{ovo}(\mathbf{\hat{b}}) = \argmin_{\mathbf{y} \in \mathcal{Y}}
(d_{ham}(\mathbf{\hat{b}}, enc_{ovo}(\mathbf{y})))$.

%% For decoding, the voting procedure in OVO is equivalent to finding the
%% nearest neighbor in the encoding space under the Hamming distance ($d_{ham}$).
%% More precisely, with the predicted encoded vector $\mathbf{\hat{b}} = h(\mathbf{x})$,
%% the decoding function is written as
%% $dec_{ovo}(\mathbf{\hat{b}}) = \argmax_{\mathbf{y} \in \mathcal{Y}}
%% (d_{ham}(\mathbf{\hat{b}}, enc_{ovo}(\mathbf{y})))$.
%% %\begin{equation} \label{eq:ovo_dec}
%% %dec_{ovo}(\mathbf{\hat{b}}) = \argmax_{\mathbf{y} \in \mathcal{Y}}
%% %(d_{ham}(\mathbf{\hat{b}}, enc_{ovo}(\mathbf{y})))
%% %\end{equation}
%% %where $d_{ham}$ represents the the Hamming distance.

%% %Next, we derive the cost-sensitive code for general cost functions from the OVO code.

The na{\"i}ve OVO for ML-ECC above suffers from several issues. First, the code length $\binom{2^K}{2}$ is prohibitively long for large $K$, making it inefficient to compute. Second, many of the $\binom{2^K}{2}$ classifiers may not be associated with enough data during training. Last but not least, OVO is not cost-sensitive and cannot adapt to different cost functions easily.
We resolve the issues in the designs below.

\subsection{Cost-sensitive encoding}
%\myparagraph{Cost-sensitive encoding.}
The OVO code is designed to optimize 0/1 loss
($C(\mathbf{y}, \hat{\mathbf{y}}) = [\![\mathbf{y} \neq \hat{\mathbf{y}}]\!]$,
where $[\![ \cdot ]\!]$ is the indicator function) for MLC.
In the OVO code, each bit of $enc_{ovo}(\mathbf{y})$ is learned from only the
instances with $\mathbf{y}$ being exactly the same as $\mathbf{y}^i_{\alpha}$
or $\mathbf{y}^i_{\beta}$.
For instances with $\mathbf{y}$ being neither $\mathbf{y}^i_{\alpha}$ nor
$\mathbf{y}^i_{\beta}$, these instances will be dropped from training.
This suits the design of optimizing 0/1 loss.
Now, we take a different perspective to view the OVO code.

When considering 0/1 loss, what the OVO code does is to decide whether
predict as $\mathbf{y}^i_{\alpha}$ or $\mathbf{y}^i_{\beta}$ suffers less 0/1
loss.
For the case that $\mathbf{y}$ is neither $\mathbf{y}^i_{\alpha}$ nor~$\mathbf{y}^i_{\beta}$,
the costs for predicting as $\mathbf{y}^i_{\alpha}$ and
$\mathbf{y}^i_{\beta}$ are the same.
That is why OVO code ignores these cases during training.
However, for other cost functions, the costs for predicting $\mathbf{y}$ as
$\mathbf{y}^i_{\alpha}$ and $\mathbf{y}^i_{\beta}$ can be different.
Hence, even if the label vector $\mathbf{y}$ is neither $\mathbf{y}^i_{\alpha}$ nor
$\mathbf{y}^i_{\beta}$, the vector can still provide information for training.

To generalize the encoding function towards cost-sensitivity, we hold the
same idea that each bit should predict which reference label vector incurs less
cost.
The encoding function is designed as
$enc_{cs}(\mathbf{y})[i] =
  \begin{cases}
    1       & \quad \text{if } C(\mathbf{y}, \mathbf{y}^i_{\alpha}) < C(\mathbf{y}, \mathbf{y}^i_{\beta}) \\
    0       & \quad \text{if } C(\mathbf{y}, \mathbf{y}^i_{\alpha}) > C(\mathbf{y}, \mathbf{y}^i_{\beta}) \\
    0.5      & \quad \text{otherwise}
  \end{cases}$.
%The cost-sensitive encoding function is designed as follows.
%\begin{equation} \label{eq:cs_enc}
%enc_{cs}(\mathbf{y})[i] =
%  \begin{cases}
%    1       & \quad \text{if } C(\mathbf{y}^i_{\alpha}, \mathbf{y}) < C(\mathbf{y}^i_{\beta}, \mathbf{y}) \\
%    0       & \quad \text{if } C(\mathbf{y}^i_{\alpha}, \mathbf{y}) > C(\mathbf{y}^i_{\beta}, \mathbf{y}) \\
%    0.5      & \quad \text{otherwise}
%  \end{cases}
%\end{equation}

\subsection{Training classifiers for cost-sensitive codes}
%\myparagraph{Training classifiers for cost-sensitive codes.}
With the encoding function defined, we learn a classifier $h$ to predict
the encoded vectors outputted from $enc_{cs}$.
Although $enc_{cs}$ gives the classifier a better ground truth, different
label vectors are not equally important for the classifier.
For example,
if $C(\mathbf{y}, \mathbf{y}^i_{\alpha})$ and
$C(\mathbf{y}, \mathbf{y}^i_{\beta})$ differ by a lot, there would
be a high cost if the classifier gives the wrong prediction, thus making
$\mathbf{y}$ very important.
In contrast, if there exists a label vector $\mathbf{y}$ s.t.
$C(\mathbf{y}, \mathbf{y}^i_{\alpha}) \approx
C(\mathbf{y}, \mathbf{y}^i_{\beta})$, then $\mathbf{y}$ is
relatively unimportant because a misclassified $\mathbf{y}$
would not incur a high cost.
Thus, we design a weight function to emphasize the importance for each
label vector as $weight(\mathbf{y})[i] = |C(\mathbf{y}, \mathbf{y}^i_{\alpha})
                                          - C(\mathbf{y}, \mathbf{y}^i_{\beta})|$.
%label vector as Equation~\ref{eq:weight}.
%\begin{equation} \label{eq:weight}
%   weight(\mathbf{y})[i] = |C(\mathbf{y}^i_{\alpha}, \mathbf{y}) - C(\mathbf{y}^i_{\beta}, \mathbf{y})|
%\end{equation}

Dataset $\{(\mathbf{x}^{(n)}, enc_{cs}(\mathbf{y}^{(n)}),
weight(\mathbf{y}^{(n)}))\}^N_{n=1}$ is used to train the classifier~$h$ to
predict the encoded vector.
Normally, $h$ should be trained on the full-length encoded vectors.
But the exponentially growing code length $\binom{2^{K}}{2}$ makes training
on the full encoding infeasible.
However, many classifiers would result in learning similar problems during
training.
This could allow us to use fewer bits and preserves the same amount of information.
For example, let the $i$-th reference label vectors be
$\mathbf{y}^i_{\alpha} = (1, 0, 1, 0)$ and
$\mathbf{y}^i_{\beta} = (1, 0, 0, 1)$, and
the $j$-th reference vectors be
$\mathbf{y}^j_{\alpha} = (1, 1, 1, 0)$ and
$\mathbf{y}^j_{\beta} = (1, 1, 0, 1)$.
The $i$-th and $j$-th classifier are actually learning similar things:
learning to predict whether the last two labels of the label vector should be
$(1, 0)$ or $(0, 1)$.
Observing the redundancy in the encoded vectors, it is clear that the length
of the encoded vector can be decreased and thus learning becomes feasible.
For simplicity, we uniformly sample some bits for from encoded vectors.
In Section \ref{experiments}, we demonstrate that the number of needed
bits are much smaller than $\binom{2^{K}}{2}$.

\subsection{Cost-sensitive decoding}
%\myparagraph{Cost-sensitive decoding.}
OVO code decodes by letting each bit votes on either of the reference label
vectors.
Following the idea for encoding, this is also a special case of decoding by
considering the 0/1 loss.
To match with our proposed cost-sensitive encoding, the decoding approach
is redesigned to utilize the information more effectively.

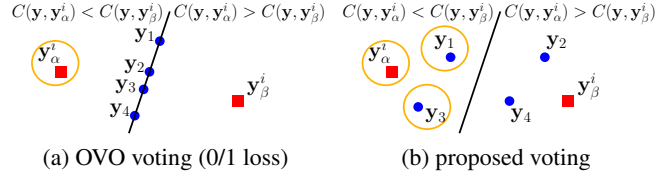
\begin{figure}[t]
\centering
\subfloat[OVO voting (0/1 loss)]{
\resizebox {.35\columnwidth} {!} {
  \begin{tikzpicture}[shorten >=1pt,draw=black!50]
      \tikzstyle{squ}=[rectangle,fill=black!25, scale=1.8];
      \tikzstyle{cir}=[circle, fill=blue, scale=1.1];
      \tikzstyle{tri}=[regular polygon, regular polygon sides=3, fill=green, scale=0.9];
      \tikzstyle{pen}=[regular polygon, regular polygon sides=5, fill=red, scale=0.9];
      \tikzstyle{hex}=[regular polygon, regular polygon sides=6, fill=orange, scale=0.9];

      \node [] (r1) at (1.6,7.7) {\huge $\mathbf{y}^i_{\alpha}$};
      \node [squ,fill=red] (r1) at (2.0,7.0) {};
      \node [] (r1) at (8.7,6.5) {\huge $\mathbf{y}^i_{\beta}$};
      \node [squ,fill=red] (r1) at (8.0,6.0) {};
      
      \draw [draw=red!30!yellow, line width=0.6mm] (1.8,7.3) ellipse (2.5em and 2.4em);

      \node [] (m1) at (4.0,5.7) {\huge $\mathbf{y}_{4}$};
      \node [cir] (m1) at (4.5,5.5) {};
      \node [] (m4) at (4.2,6.5) {\huge $\mathbf{y}_{3}$};
      \node [cir] (m4) at (4.8,6.4) {};
      \node [] (m2) at (4.5,7.2) {\huge $\mathbf{y}_{2}$};
      \node [cir] (m2) at (5.0,7.0) {};
      \node [] (m3) at (4.85,8.25) {\huge $\mathbf{y}_{1}$};
      \node [cir] (m3) at (5.35,8.05) {};

      \node [] (m3) at (2.9,9) { \LARGE
        $C(\mathbf{y}, \mathbf{y}^i_{\alpha}) < C(\mathbf{y}, \mathbf{y}^i_{\beta})$};
      \node [] (m3) at (8.4,9) { \LARGE
        $C(\mathbf{y}, \mathbf{y}^i_{\alpha}) > C(\mathbf{y}, \mathbf{y}^i_{\beta})$};

      \draw[black,line width=0.6mm] (4.3,4.9) -- (5.7,9.1);
  \end{tikzpicture}
}}
\subfloat[proposed voting]{
\resizebox {.35\columnwidth} {!} {
  \begin{tikzpicture}[shorten >=1pt,draw=black!50]
      \tikzstyle{squ}=[rectangle,fill=black!25, scale=1.8];
      \tikzstyle{cir}=[circle, fill=blue, scale=1.1];
      \tikzstyle{tri}=[regular polygon, regular polygon sides=3, fill=green, scale=0.9];
      \tikzstyle{pen}=[regular polygon, regular polygon sides=5, fill=red, scale=0.9];
      \tikzstyle{hex}=[regular polygon, regular polygon sides=6, fill=orange, scale=0.9];

      \node [] (r1) at (1.6,7.7) {\huge $\mathbf{y}^i_{\alpha}$};
      \node [squ,fill=red] (r1) at (2.0,7.0) {};
      \node [] (r1) at (8.7,6.5) {\huge $\mathbf{y}^i_{\beta}$};
      \node [squ,fill=red] (r1) at (8.0,6.0) {};
      
      \draw [draw=red!30!yellow, line width=0.6mm] (1.8,7.3) ellipse (2.5em and 2.4em);
      \draw [draw=red!30!yellow, line width=0.6mm] (3.8,7.8) ellipse (2.5em and 2.4em);
      \draw [draw=red!30!yellow, line width=0.6mm] (3.2,5.8) ellipse (2.5em and 2.4em);

      %\node [cir] (r1) at (4.0,4.0) {};
      \node [] (m1) at (6.4,5.5) {\huge $\mathbf{y}_{4}$};
      \node [cir] (m1) at (6.0,6.0) {};
      \node [] (m4) at (3.5,5.5) {\huge $\mathbf{y}_{3}$};
      \node [cir] (m4) at (2.9,5.8) {};
      \node [] (m2) at (7.6,8.0) {\huge $\mathbf{y}_{2}$};
      \node [cir] (m2) at (7.2,7.5) {};
      \node [] (m3) at (3.8,8.0) {\huge $\mathbf{y}_{1}$};
      \node [cir] (m3) at (4.0,7.5) {};
      %\node [cir] (r1) at (6.0,10.0) {};

      \node [] (m3) at (2.9,9) { \LARGE
        $C(\mathbf{y}, \mathbf{y}^i_{\alpha}) < C(\mathbf{y}, \mathbf{y}^i_{\beta})$};
      %\node [] (m3) at (5.5,9.5) {
      %  $C(\mathbf{y}, \mathbf{y}^i_{\alpha}) = C(\mathbf{y}, \mathbf{y}^i_{\beta})$};
      \node [] (m3) at (8.4,9) { \LARGE
        $C(\mathbf{y}, \mathbf{y}^i_{\alpha}) > C(\mathbf{y}, \mathbf{y}^i_{\beta})$};

      %\draw[black,very thick] (4.5,5.5) -- (5.6,8.8);
      \draw[black,line width=0.6mm] (4.3,4.9) -- (5.7,9.1);
  \end{tikzpicture}
}}
\caption{An illustration of the decoding methods.
} \label{fig:decode-example}
\end{figure}

Figure \ref{fig:decode-example} is an illustration of the relation between
encoded vectors under OVO encoding and our cost-sensitive encoding.
In 0/1 loss, all instances that are predicted incorrectly incur the same
cost making all label vectors except~$\mathbf{y}^i_{\alpha}$ and~$\mathbf{y}^i_{\beta}$ on the decision boundary.
Only $\mathbf{y}^i_{\alpha}$ and $\mathbf{y}^i_{\beta}$ are distinguishable
under the current bit.
Thus, original OVO voting only needs to be done on reference label vectors.
When using our cost-sensitive encoding, all label vectors are generally
separated into two groups by the boundary as Figure
\ref{fig:decode-example}(b):
the group that is closer to $\mathbf{y}^i_{\alpha}$ (left) (in terms of cost)
and the group that is closer to $\mathbf{y}^i_{\beta}$.
A predicted encoded bit not only provides the information about the reference
label vector, but also the information about all other label vectors in the
same group.
Following this thought, if the prediction is $\mathbf{y}^i_{\alpha}$, all label vectors~$\mathbf{y}$
such that $C(\mathbf{y}, \mathbf{y}^i_{\alpha}) <
C(\mathbf{y}, \mathbf{y}^i_{\beta})$ should be voted as well.
If predicted otherwise, all label vectors in the other group are voted.
By this voting approach, we can use the information encoded within the vectors to
decode more effectively.

In fact, this voting approach echoes the Hamming decoding 
for ECC~\cite{allwein2001reducing}.
More specifically, with the predicted encoded vector $\mathbf{\hat{b}} =
h(\mathbf{x})$, the decoding function is written as
$dec_{cs}(\mathbf{\hat{b}}) = \argmin_{\mathbf{y} \in \mathcal{Y}}
d_{ham}(\mathbf{\hat{b}}, enc_{cs}(\mathbf{y}))$.
With this formulation, $dec_{cs}$ is formulated as the classic
nearest neighbor search problem, where efficient algorithms exist
to speed up the decoding process \cite{liu2006new}.
%\cite{bentley1975multidimensional,liu2006new}.

Despite the efficient decoding algorithm, the number of possible predictions $|\mathcal{Y}|$ equals $2^K$, which makes it computationally infeasible.
%to search through all possible label vectors.
%To solve this problem, we propose only working with a subset of label vectors
%that are more likely to be the prediction.
Inspired by \cite{KH2017}, we propose to only work with a subset of label vectors
that are more likely to be the prediction.
We define a relevant set
$\mathcal{\tilde{Y}} \subseteq \mathcal{Y}$, which contains a subset of
the label vectors from the label space, on which we perform the nearest neighbor search.
The decoding function is written as
$dec_{cs}(\mathbf{\hat{b}}) = \argmin_{\mathbf{y} \in \mathcal{\tilde{Y}}} d_{ham}(\mathbf{\hat{b}}, enc_{cs}(\mathbf{y}))$.
%Thus, $dec_{cs}$ is rewritten as follows.
%\begin{equation} \label{eq:cs_dec}
%	dec_{cs}(\mathbf{\hat{b}}) = \argmax_{\mathbf{y} \in \mathcal{\tilde{Y}}} d_{ham}(\mathbf{\hat{b}}, enc_{cs}(\mathbf{y}))
%\end{equation}

The use of the $\mathcal{\tilde{Y}}$ introduces a trade-off between
the number of possible predictions and the prediction efficiency.
A reasonable choice of $\mathcal{\tilde{Y}}$ would be
$\{\mathbf{y} | (\mathbf{x}, \mathbf{y}) \in \mathcal{D}\}$,
which are the distinct label vectors in the training set.
Given that the training and testing sets come from the same distribution, the
label vectors that appear in the testing set are likely to have appeared in
the training set.
%Even if some label vectors do not appear in the training set but appear in the
%testing set, the label vector with the smallest cost would be predicted.
We justify this choice of $\mathcal{\tilde{Y}}$ in Section \ref{experiments}.

The algorithm that combines $enc_{cs}$, $weight$ and $dec_{cs}$ is
called \textit{cost-sensitive reference pair encoding} (CSRPE).
Our design is inspired by a cost-sensitive extension of OVO for MCC problem
called \textit{cost-sensitive one-versus-one}~\cite{DBLP:conf/acml/Lin14}, but
is refined by our special ideas for encoding and
decoding in the MLC problem.

\section{Active Learning for CSMLC} \label{csmlal}

CSRPE is able to preserve cost information in the encoded vectors.
In this section, we design a novel active learning algorithm for MLC based on CSRPE.

MLC algorithms intend to learn a classifier from a fully labeled dataset, in
which every feature vector is paired with a label vector.
In many real-world applications, obtaining a label vector to the corresponding
feature vector is very expensive \cite{liu2004active}.
This gives rise to a new problem, active learning, which investigates how
to obtain good performance with as little data labeled as possible.

In this paper, we consider the pool-based \textit{multi-label active learning}
(MLAL) setting \cite{settles2010active} and formulate the cost-sensitive
extension of MLAL called
\textit{cost-sensitive multi-label active learning}~(CSMLAL).
In CSMLAL, the algorithm is presented with two sets of data,
the labeled pool
$\mathcal{D}_l = \{(\mathbf{x}^{(n)}, \mathbf{y}^{(n)})\}^{N_l}_{n=1}$
and the unlabeled pool
$\mathcal{D}_u = \{\mathbf{x}^{(n)}\}^{N_u}_{n=1}$.
During iterations $t = 1, \ldots, T$, the MLAL algorithm considers
$\mathcal{D}_u$, $\mathcal{D}_l$, a MLC classifier $f_t$ trained on
$\mathcal{D}_l$ and cost function $C$
to choose a instance $\mathbf{x}_t \in \mathcal{D}_u$ to query.
After the queried label vector is retrieved as $\mathbf{y}_t$,
$\mathbf{x}_t$ is removed from $\mathcal{D}_u$ and the pair
$(\mathbf{x}_t, \mathbf{y}_t)$ is added to $\mathcal{D}_l$.
With a small budget of $T$ queries, the goal of the CSMLAL algorithm is
to minimize the average prediction cost of $f_t$ on the testing instances
evaluated on $C$.

%Apart from the above setting, there is another setting of MLAL
%that queries each bit on each instance instead of querying each label
%vector \cite{huang2013active,huang2015multi,DBLP:journals/pami/HuangJZ14}.
%In this paper, we focus on the above setting of querying the whole label
%vector.

Many of the current MLAL algorithms are based on the idea of
\textit{uncertainty sampling}.
They query the instance that current classifier $f_t$ is most uncertain
about.
There are different uncertainty measures being developed.
However, most of these measures consider only one specific $C$ or even
completely ignoring $C$.
\textit{Binary minimization} \cite{brinker2006active} was proposed
to directly take the most uncertain bit in the label vector to represent the
uncertainty of the whole instance.
It queries based on one label at a time and arguably optimizes towards Hamming loss.
Another work, in contrast, calculates an average over the
uncertainty of all labels \cite{yang2009effective}.
%\textit{Maximal loss reduction with maximal confidence}
%\cite{yang2009effective}, in contrast, calculates an average over the
%uncertainty of all labels.
%It tries to do well on the whole label vector which is more aligned with 0/1 loss.
%Another MLAL algorithm called \textit{adaptive active learning}
%\cite{li2013active} uses the difference between the most
%uncertain relevant label and irrelevant label as an uncertainty measure.
Yet another work uses the difference between the most uncertain relevant label
and irrelevant label as an uncertainty measure \cite{li2013active}.
This uncertainty is then combined with label cardinality inconsistency.
However, this measure is designed heuristically and does not aim at any
$C$.

We propose \textit{cost-sensitive uncertainty} in the encoded vector
space to evaluate the importance of instances.
The cost-sensitive uncertainty can be separated into two parts, the
\textit{cost estimation uncertainty} and the \textit{cost utility
uncertainty}.

\subsection{Cost estimation uncertainty}
%\myparagraph{Cost estimation uncertainty.}
Cost estimation uncertainty measures how well CSRPE estimates the cost between
label vectors.
Let the predicted encoded vector $\mathbf{\hat{b}} = h(\mathbf{x})$ and
$\mathbf{\tilde{b}} = enc_{cs}(dec_{cs}(\mathbf{\hat{b}}))$.
Note that $\mathbf{\tilde{b}}$ is actually the nearest encoded vector of $\mathbf{\hat{b}}$.
Ideally, if CSRPE estimates the cost information well, $\mathbf{\hat{b}}$
should be close to $\mathbf{\tilde{b}}$.
If, unfortunately, the distance $d_{ham}(\mathbf{\hat{b}},
\mathbf{\tilde{b}})$ is large, this implies that CSRPE does not have a good cost
estimation for this $\mathbf{x}$ and we hence need more information about it.
In other words, we are uncertain about this $\mathbf{x}$.
For this reason, we define $d_{ham}(\mathbf{\hat{b}}, \mathbf{\tilde{b}})$ as
the \textit{cost estimation uncertainty}.

\subsection{Cost utility uncertainty.}
%\myparagraph{Cost utility uncertainty.}
The cost utility uncertainty measures how uncertain the classifier $f_t$ is under
the current cost function.
Let the prediction $\mathbf{\bar{y}} = f_t(\mathbf{x})$ and
its encoding $\mathbf{\bar{b}} = enc_{cs}(\mathbf{\bar{y}})$.
If the classifier $f_t$ is certain about its prediction under current cost
function, $\mathbf{\bar{b}}$ should be close to the cost estimation
$\mathbf{\hat{b}} = h(\mathbf{x})$.
If unfortunately, distance $d_{ham}(\mathbf{\hat{b}}, \mathbf{\bar{b}})$
is large, it implies that classifier $f_t$ is uncertain under the current cost
function.
Therefore, we define $d_{ham}(\mathbf{\hat{b}}, \mathbf{\bar{b}})$ as the
\textit{cost utility uncertainty}.

The proposed cost-sensitive uncertainty is the combination of these two parts of uncertainty, namely
$d_{ham}(\mathbf{\hat{b}}, \mathbf{\tilde{b}}) + d_{ham}(\mathbf{\hat{b}},
\mathbf{\bar{b}})$.
The cost-sensitive uncertainty leads to a novel algorithm for
CSMLAL.
For each iteration, the algorithm selects the instance with the highest
cost-sensitive uncertainty to query its label.

\section{Experiments}\label{experiments}

We justify the proposed algorithm on ten public datasets \cite{mulan} and
four evaluation criteria.
%, including F1 score, Accuracy score, Hamming loss, and Rank loss.
The dataset statistics are listed in Table \ref{tab:dataset_stats} and the
definition of the evaluation criteria are
$\text{F1 score}(\mathbf{y}, \mathbf{\hat{y}}) = \frac{2\| \mathbf{y} \cap
\mathbf{\hat{y}}\|_1}{\|\mathbf{y}\|_1 + \|\mathbf{\hat{y}}\|_1},
\text{Accuracy score}(\mathbf{y}, \mathbf{\hat{y}}) = \frac{\|\mathbf{y} \cap \mathbf{\hat{y}}\|_1}{\|\mathbf{y} \cup \mathbf{\hat{y}}\|_1},
\text{Hamming loss}(\mathbf{y}, \mathbf{\hat{y}}) = \frac{1}{K} \sum_{k=1}^K
[\![\mathbf{y}[k] \neq \mathbf{\hat{y}}[k]]\!]$
and
$\text{Rank loss}(\mathbf{y}, \mathbf{\hat{y}}) = \sum_{\mathbf{y}[i]>\mathbf{y}[j]}\left([\![\mathbf{\hat{y}}[i]<\mathbf{\hat{y}}[j]]\!] +
\frac{1}{2}[\![\mathbf{\hat{y}}[i]=\mathbf{\hat{y}}[j]]\!]\right)$ \cite{DBLP:reference/dmkdh/TsoumakasKV10}.
$[\![ \cdot ]\!]$ represents the indicator function.
The experiment was run 20 times, each with a random 50-50 training-testing
split.
CSRPE has the flexibility to take any base learner.
In CSMLC experiments, CSRPE is viewed as an ensemble MLC method, each
bit with a binary classifier attached.
Because ensemble of decision trees is arguably a popular ensemble
method nowadays, we use decision trees as the base learner in these experiments.
The parameters are searched with 3-fold cross-validation.

\begin{table}[H]
\centering
\caption{Dataset statistics}
\label{tab:dataset_stats}
\scriptsize
\vspace{1em}
\begin{tabularx}{.5\columnwidth}{@{} *6{>{\centering\arraybackslash}X}@{}} \hline
Dataset & labels & instances & features & density & distinct \\ \hline
Corel5k & 374 & 5000 & 499 & 0.009 & 3175 \\
CAL500 & 174 & 502 & 68 & 0.150 & 502 \\
bibtex & 159 & 7395 & 1836 & 0.015 & 2856 \\
enron & 53 & 1702 & 1001 & 0.064 & 753 \\
medical & 45 & 978 & 1449 & 0.028 & 94 \\
genbase & 27 & 662 & 1186 & 0.046 & 32 \\
yeast & 14 & 2417 & 103 & 0.303 & 198 \\
flags & 7 & 194 & 19 & 0.485 & 54 \\
scene & 6 & 2407 & 294 & 0.179 & 15 \\
emotions & 6 & 593 & 72 & 0.311 & 27 \\ \hline
\end{tabularx}
\end{table}

In CSMLAL experiments, the experiments are repeated for 10 runs.
Since many competitors designed their algorithms based on linear base
learners, the base learner is changed to logistic regression for fair
comparison.
The parameters are searched with 5-fold cross-validation using the
initial dataset.

%More detailed experimental setup can be found in the full version \cite{yang2016cost}.
In the following experimental results, we use $\uparrow$ ($\downarrow$) to
indicate that a higher (lower) value for the criterion is better.

\subsection{Effect of Code Length}
%\myparagraph{Effect of Code Length.}
%In Section \ref{proposed}, we mentioned that the encoded bits are sharing
%similar information, which allows the code length to be reduced by sampling.
%We justify this statement by analyzing the performance of CSRPE with respect
%to the code length.
To justify our claim in Section \ref{proposed} that the code length can be
reduced by sampling, we conduct experiments to analyzing the performance of
CSRPE with respect to the code length.

\begin{figure}[H]
\centering
\subfloat[Corel5k]{\includegraphics[width=.21\textwidth]{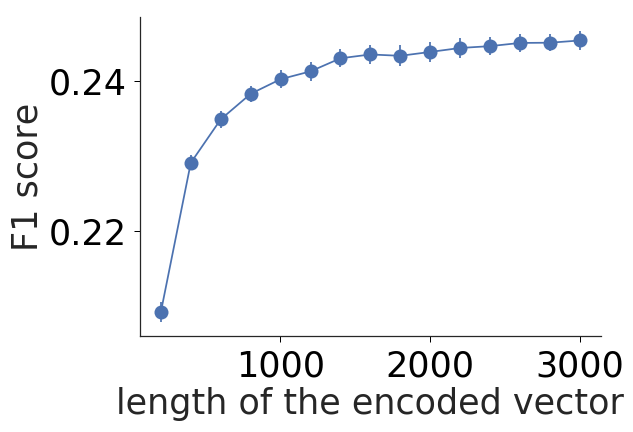}}
\hspace{1em}
\subfloat[CAL500]{\includegraphics[width=.21\textwidth]{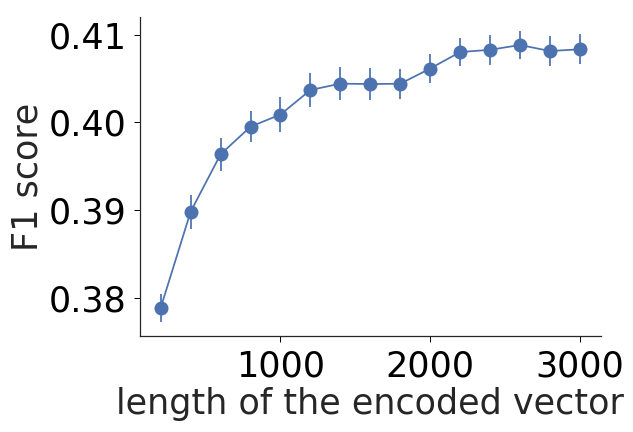}}
\hspace{1em}
\subfloat[bibtex]{\includegraphics[width=.21\textwidth]{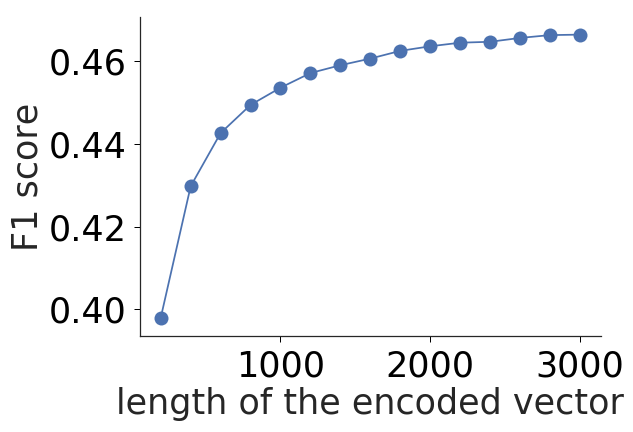}}
\hspace{1em}
\subfloat[enron]{\includegraphics[width=.21\textwidth]{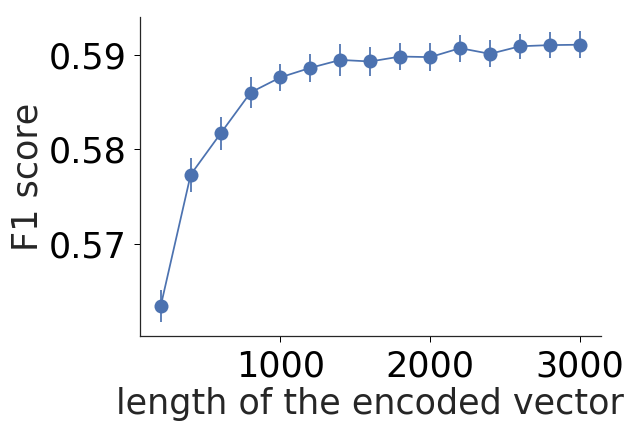}}

\subfloat[medical]{\includegraphics[width=.21\textwidth]{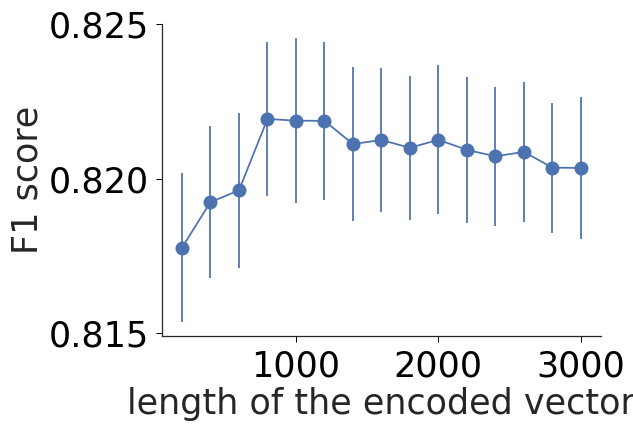}}
\hspace{1em}
\subfloat[genbase]{\includegraphics[width=.21\textwidth]{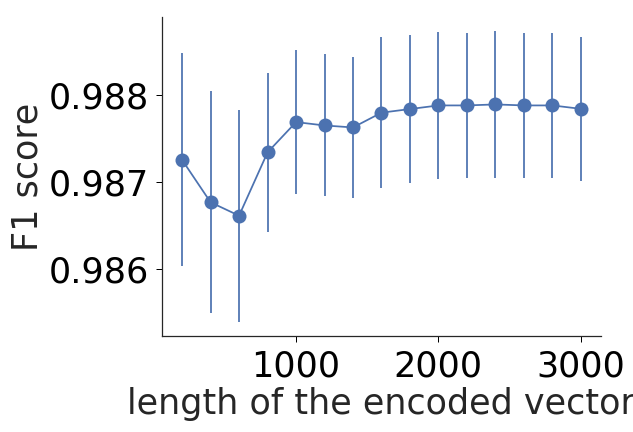}}
\hspace{1em}
\subfloat[yeast]{\includegraphics[width=.21\textwidth]{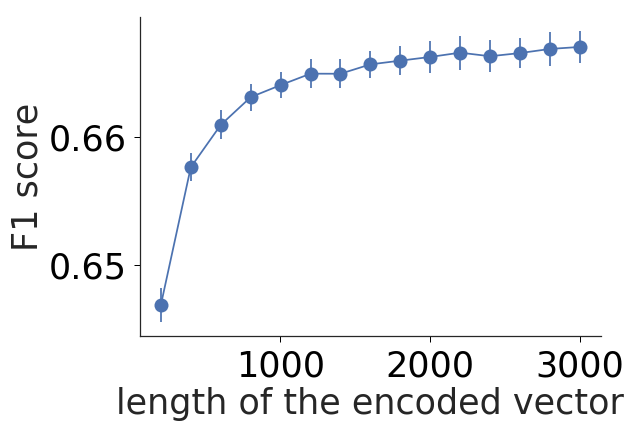}}
\hspace{1em}
\subfloat[flags]{\includegraphics[width=.21\textwidth]{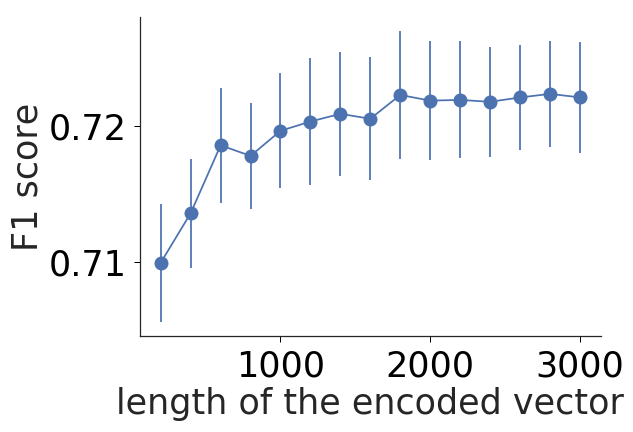}}

\subfloat[scene]{\includegraphics[width=.21\textwidth]{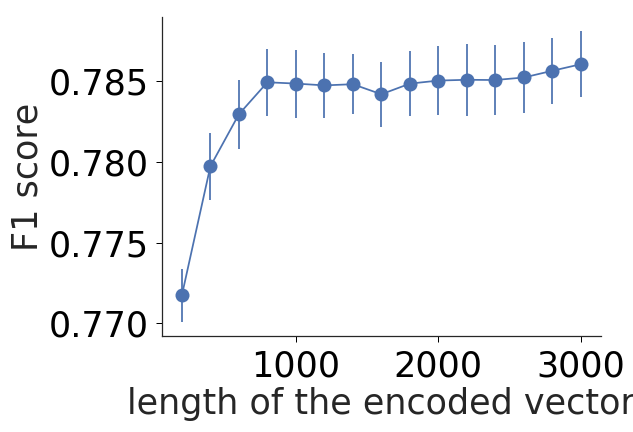}}
\hspace{1em}
\subfloat[emotions]{\includegraphics[width=.21\textwidth]{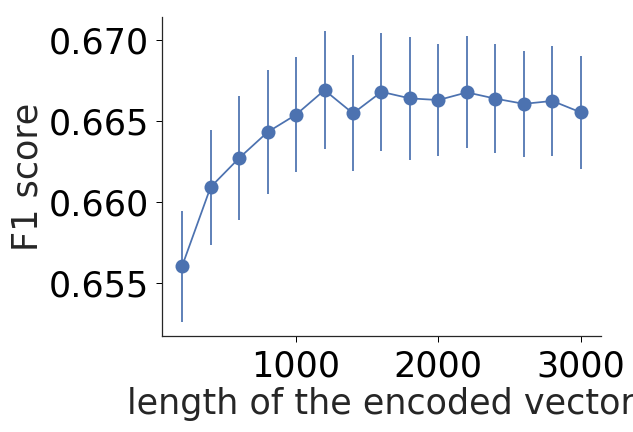}}
\caption{F1 score $\uparrow$ versus code length for CSRPE}
\label{fig:converge_f1}
\end{figure}

\begin{figure}[H]
\centering
\subfloat[Corel5k]{\includegraphics[width=.21\textwidth]{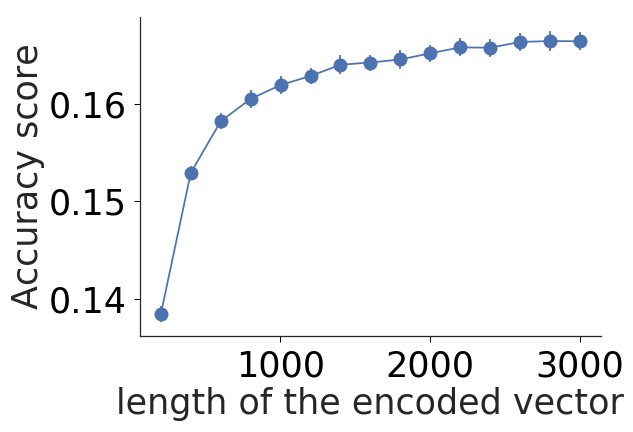}}
\hspace{1em}
\subfloat[CAL500]{\includegraphics[width=.21\textwidth]{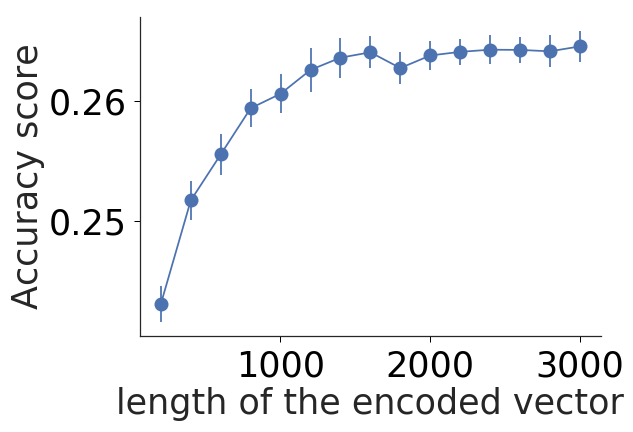}}
\hspace{1em}
\subfloat[bibtex]{\includegraphics[width=.21\textwidth]{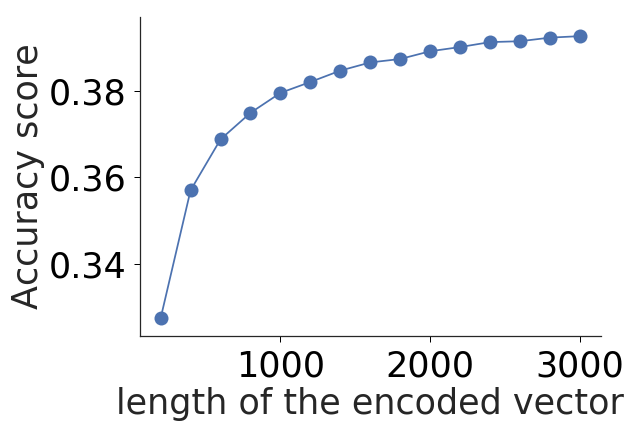}}
\hspace{1em}
\subfloat[enron]{\includegraphics[width=.21\textwidth]{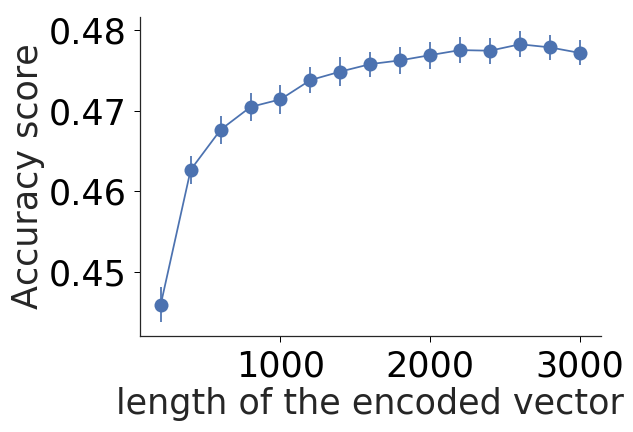}}

\subfloat[medical]{\includegraphics[width=.21\textwidth]{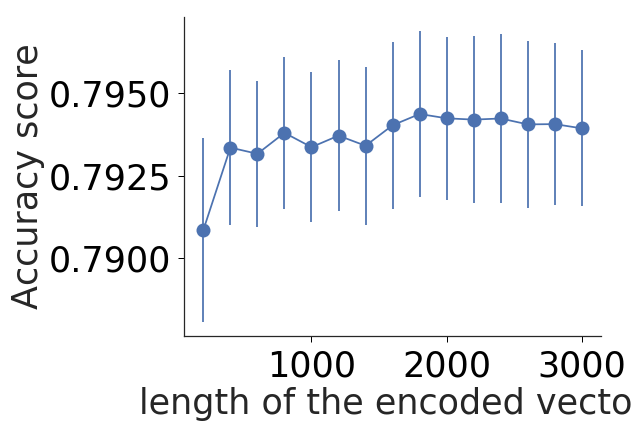}}
\hspace{1em}
\subfloat[genbase]{\includegraphics[width=.21\textwidth]{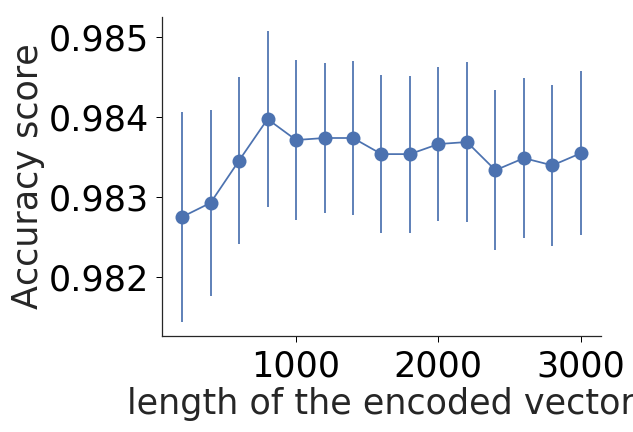}}
\hspace{1em}
\subfloat[yeast]{\includegraphics[width=.21\textwidth]{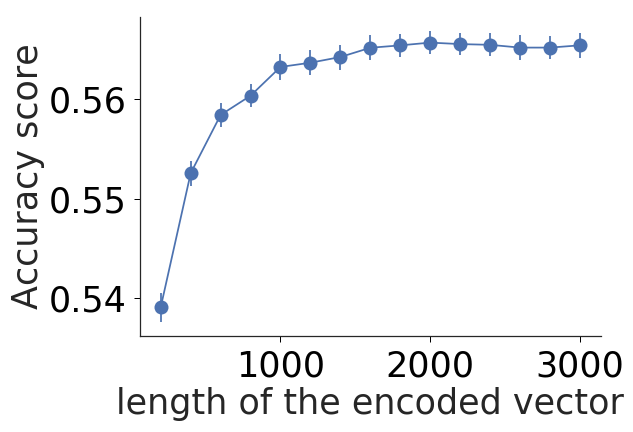}}
\hspace{1em}
\subfloat[flags]{\includegraphics[width=.21\textwidth]{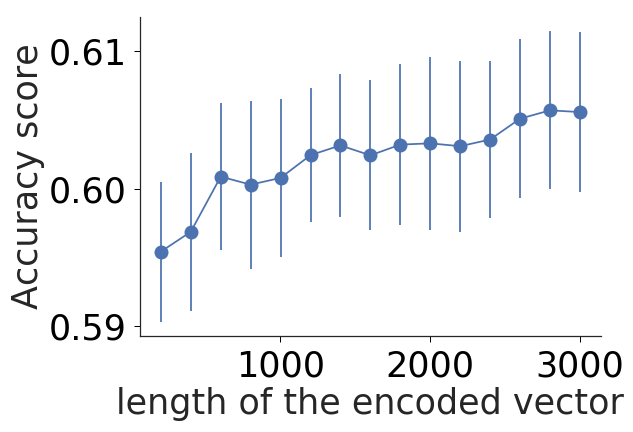}}

\subfloat[scene]{\includegraphics[width=.21\textwidth]{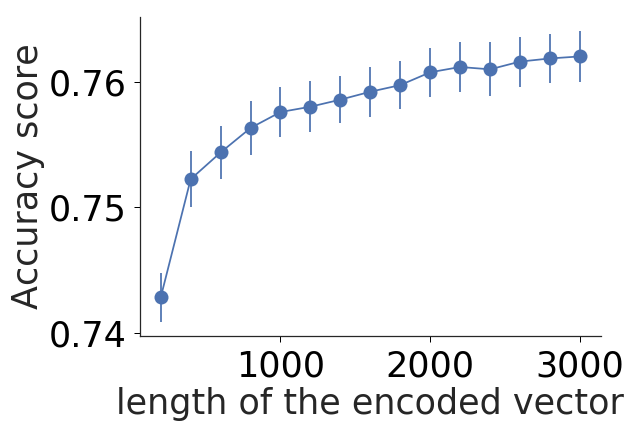}}
\hspace{1em}
\subfloat[emotions]{\includegraphics[width=.21\textwidth]{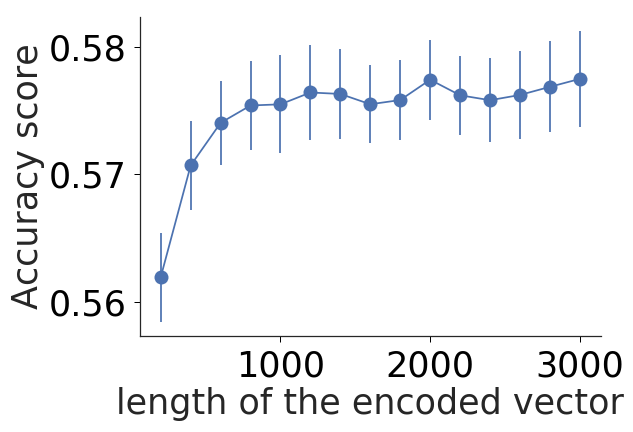}}
\caption{Accuracy score $\uparrow$ versus code length for CSRPE}
\label{fig:converge_acc}
\end{figure}

\begin{figure}[H]
\centering
\subfloat[Corel5k]{\includegraphics[width=.21\textwidth]{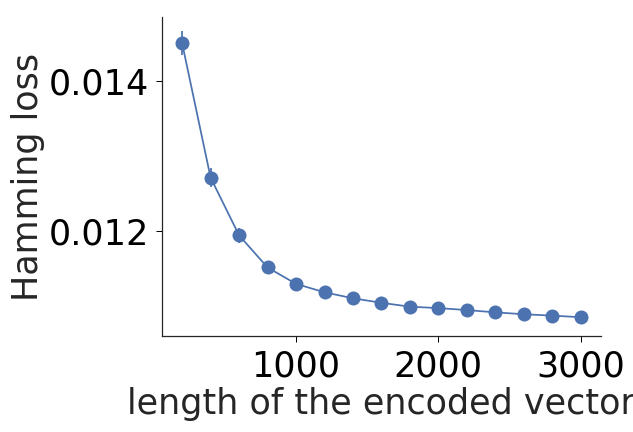}}
\hspace{1em}
\subfloat[CAL500]{\includegraphics[width=.21\textwidth]{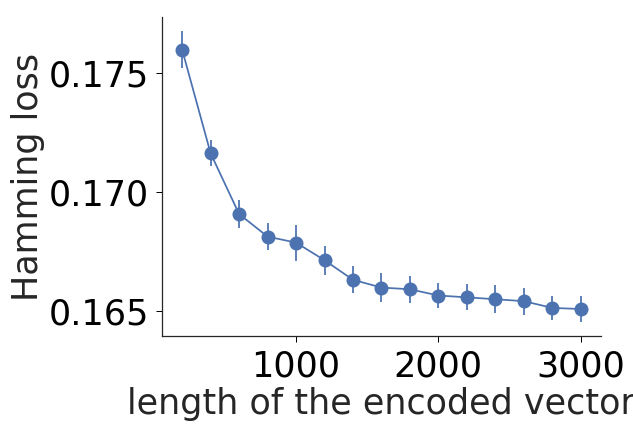}}
\hspace{1em}
\subfloat[bibtex]{\includegraphics[width=.21\textwidth]{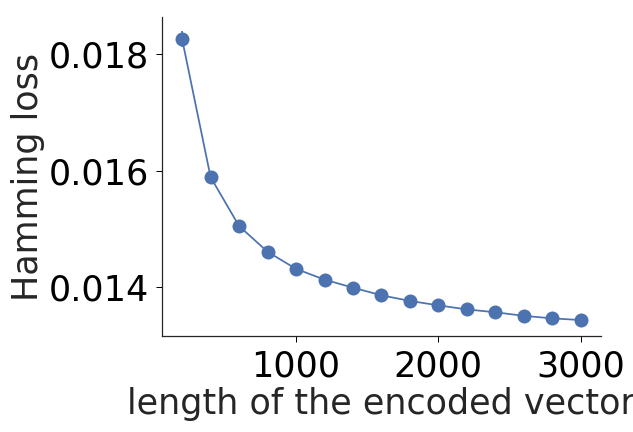}}
\hspace{1em}
\subfloat[enron]{\includegraphics[width=.21\textwidth]{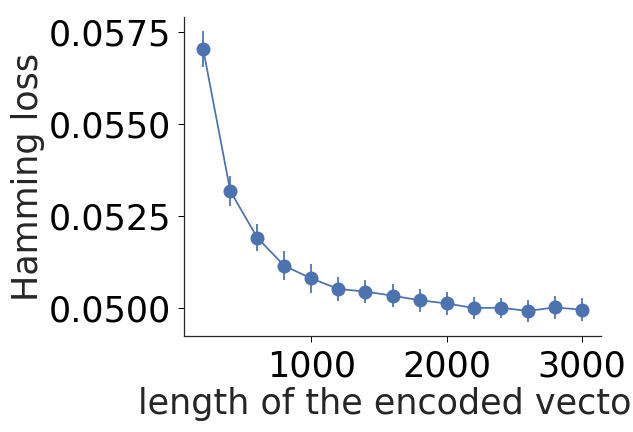}}

\subfloat[medical]{\includegraphics[width=.21\textwidth]{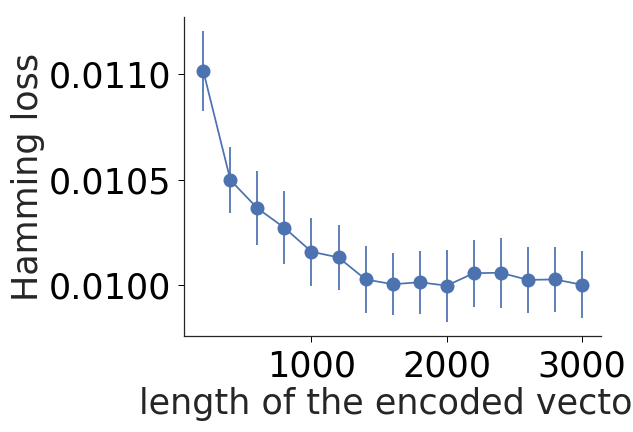}}
\hspace{1em}
\subfloat[genbase]{\includegraphics[width=.21\textwidth]{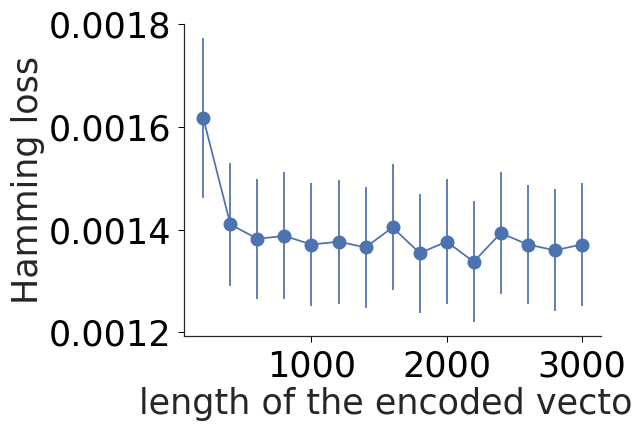}}
\hspace{1em}
\subfloat[yeast]{\includegraphics[width=.21\textwidth]{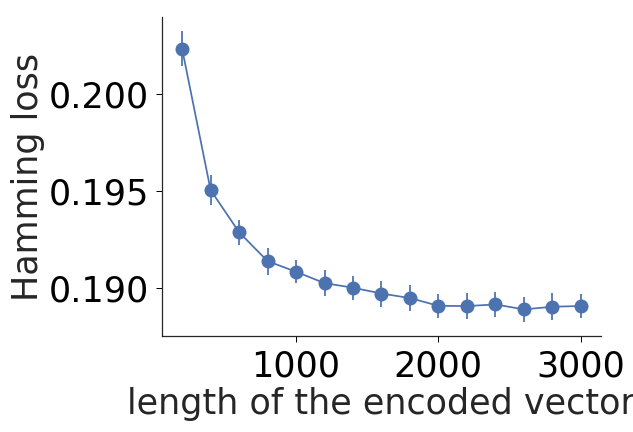}}
\hspace{1em}
\subfloat[flags]{\includegraphics[width=.21\textwidth]{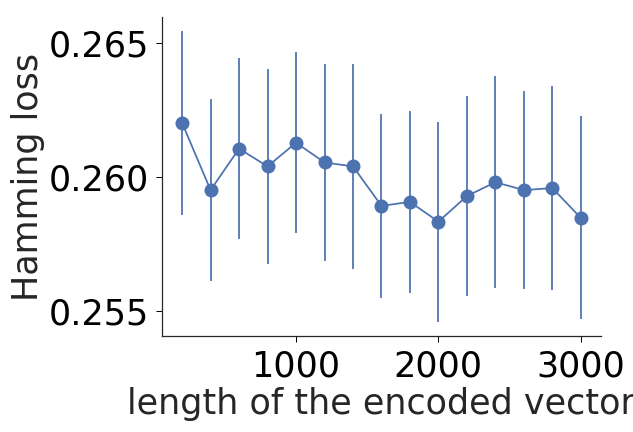}}

\subfloat[scene]{\includegraphics[width=.21\textwidth]{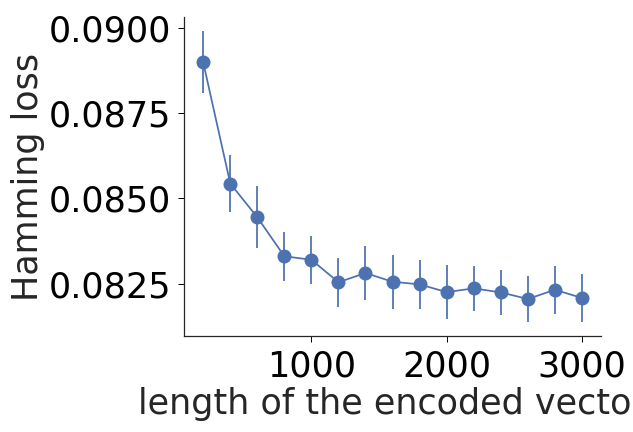}}
\hspace{1em}
\subfloat[emotions]{\includegraphics[width=.21\textwidth]{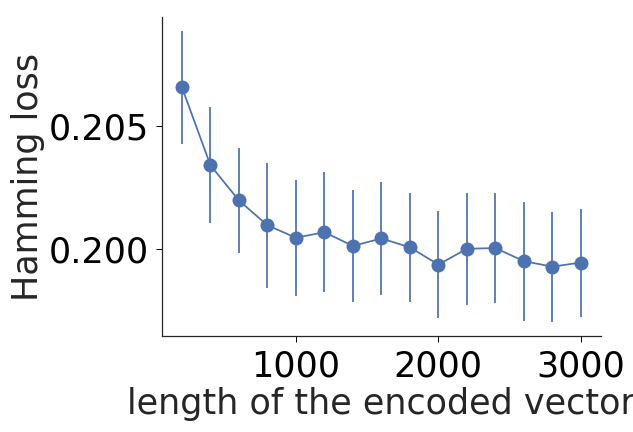}}
\caption{Hamming loss $\downarrow$ versus code length for CSRPE}
\label{fig:converge_hamming}
\end{figure}

\begin{figure}[H]
\centering
\subfloat[Corel5k]{\includegraphics[width=.21\textwidth]{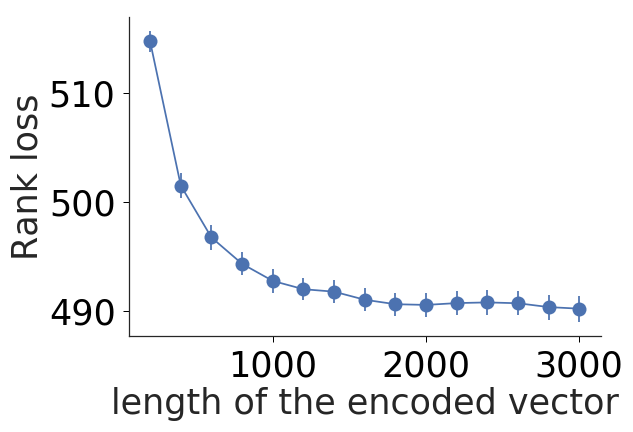}}
\hspace{1em}
\subfloat[CAL500]{\includegraphics[width=.21\textwidth]{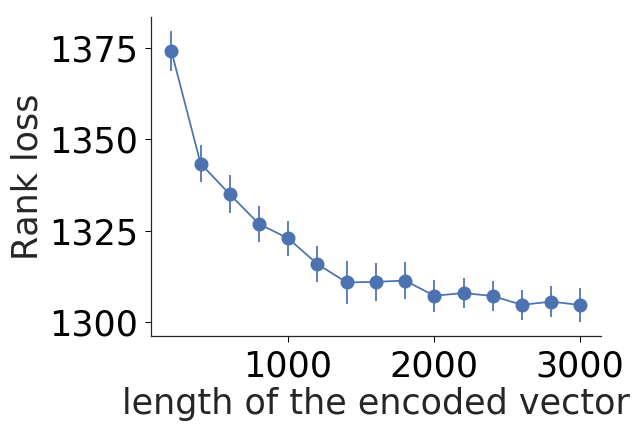}}
\hspace{1em}
\subfloat[bibtex]{\includegraphics[width=.21\textwidth]{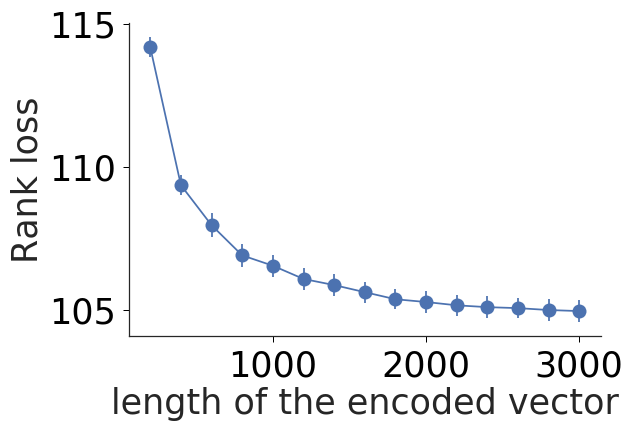}}
\hspace{1em}
\subfloat[enron]{\includegraphics[width=.21\textwidth]{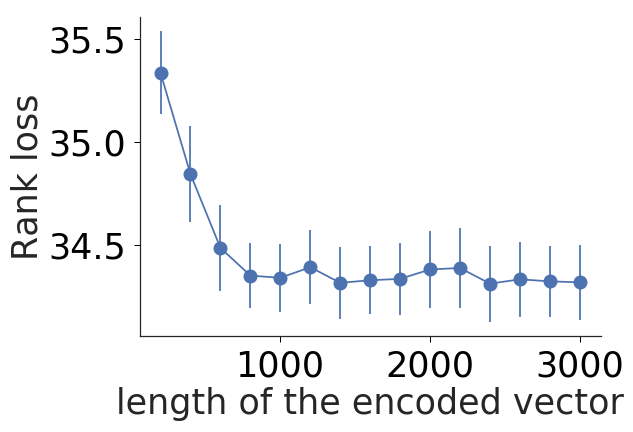}}

\subfloat[medical]{\includegraphics[width=.21\textwidth]{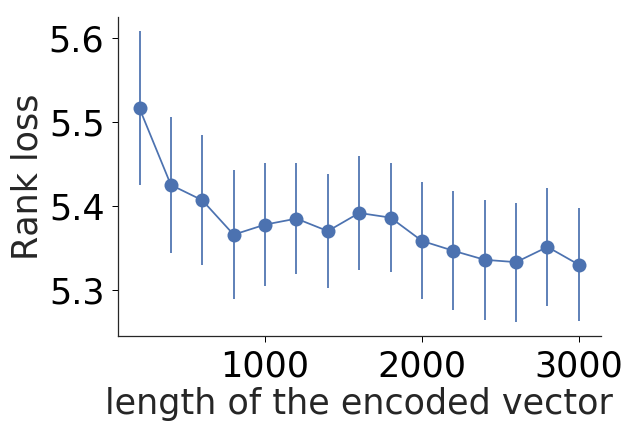}}
\hspace{1em}
\subfloat[genbase]{\includegraphics[width=.21\textwidth]{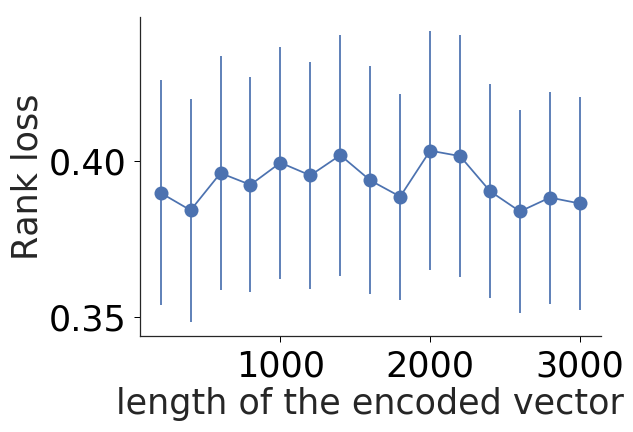}}
\hspace{1em}
\subfloat[yeast]{\includegraphics[width=.21\textwidth]{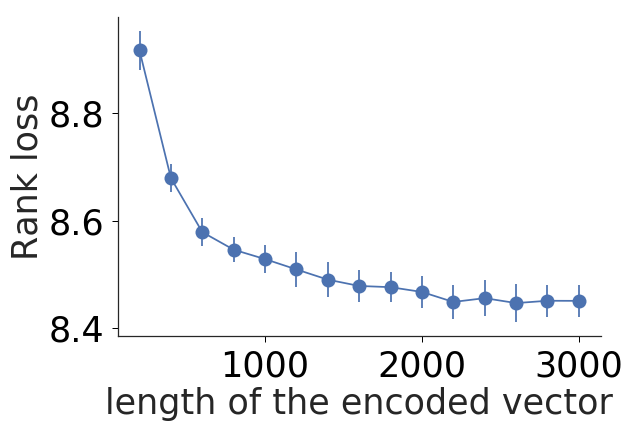}}
\hspace{1em}
\subfloat[flags]{\includegraphics[width=.21\textwidth]{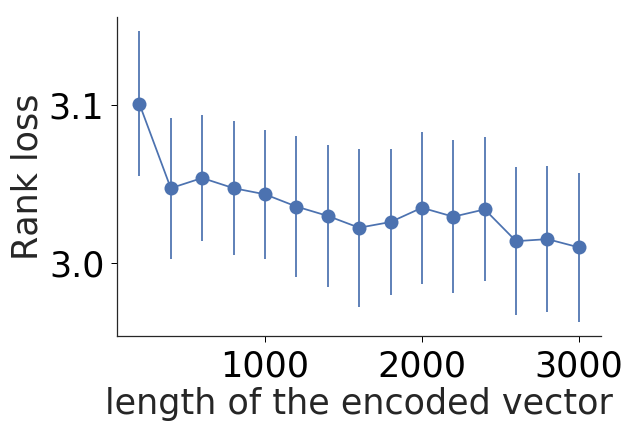}}

\subfloat[scene]{\includegraphics[width=.21\textwidth]{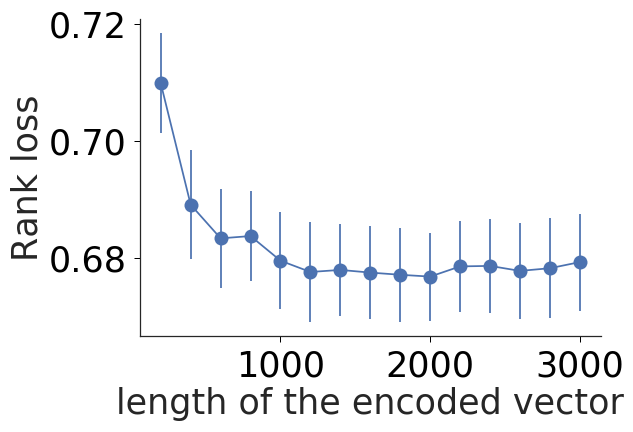}}
\hspace{1em}
\subfloat[emotions]{\includegraphics[width=.21\textwidth]{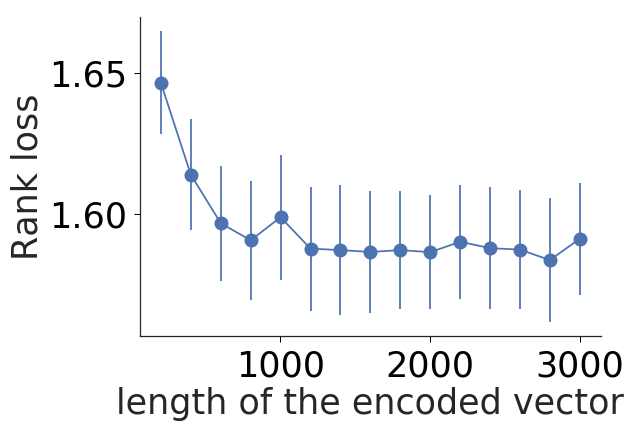}}
\caption{Rank loss $\downarrow$ versus code length for CSRPE}
\label{fig:converge_rank}
\end{figure}

Figures \ref{fig:converge_f1}, \ref{fig:converge_acc}, \ref{fig:converge_hamming} and \ref{fig:converge_rank}
%Figure \ref{fig:converge}
show the average performance and standard error versus code length.
We select two of the datasets with larger label counts to showcase the
effect of the code length on performance.
%The results of other datasets can be found in \cite{yang2016cost}.
From the figure, CSRPE performs better as the number of bit increases.
The performance of CSRPE generally converges when the code length 
reaches $3000$ across all cost functions and datasets.
The length is significantly smaller than the full encoding ($2^K$).
This justifies our claim that full encoding is not needed to achieve top
performance. % because of the redundant information in the encoding.
In the following experiments, we set the code length as $3000$.

\subsection{Influence of the Relevant Set}
%\myparagraph{Influence of the Relevant Set.}
In Section \ref{proposed}, we claim that a good choice for relevant set
$\mathcal{\tilde{Y}}$ is all distinct label vectors in the training dataset.
To justify our claim, we demonstrate that the possible downside of this
choice, which is the inability to predict all possible label vectors, will
not degrade the performance much. In particular, we compare CSRPE with CSRPE-ext,
which is CSRPE-ext with a larger relevant set that includes label vectors that
appeared in either the training set or the testing set.

\begin{table*}[ht!]
\centering
%\caption{Experimental results (mean $\pm$ ste) of CSRPE versus CSRPE-extend}
\caption{Experiment results (mean $\pm$ ste) of CSRPE and CSRPE-ext (best in bold)}
\label{tab:perfect}
\tiny
\begin{tabularx}{1\columnwidth}{*5{>{\centering\arraybackslash}X}} \hline
\multicolumn{1}{c}{Dataset}
 & \multicolumn{2}{c}{F1 score $\uparrow$} & \multicolumn{2}{c}{Accuracy score $\uparrow$} \\ \hline
 & CSRPE & CSRPE-ext & CSRPE & CSRPE-ext \\ \hline
Corel5k
& $\mathbf{0.2455 \pm 0.0012}$ & $0.2492 \pm 0.0011$
& $0.1664 \pm 0.0009$ & $\mathbf{0.1674 \pm 0.0009}$
\\
bibtex
& $0.4663 \pm 0.0008$ & $\mathbf{0.4695 \pm 0.0009}$
& $0.3926 \pm 0.0011$ & $\mathbf{0.3946 \pm 0.0010}$  \\
CAL500
& $0.4083 \pm 0.0017$ & $\mathbf{0.4109 \pm 0.0013}$
& $0.2645 \pm 0.0013$ & $\mathbf{0.2690 \pm 0.0014}$
\\
enron
& $0.5911 \pm 0.0014$ & $\mathbf{0.5921 \pm 0.0016}$
& $0.4772 \pm 0.0016$ & $\mathbf{0.4777 \pm 0.0017}$
\\
medical
& $0.8203 \pm 0.0023$ & $\mathbf{0.8204 \pm 0.0023}$
& $\mathbf{0.7939 \pm 0.0024}$ & $0.7934 \pm 0.0022$
\\
genbase
& $\mathbf{0.9878 \pm 0.0009}$ & $0.9876 \pm 0.0009$
& $\mathbf{0.9836 \pm 0.0010}$ & $0.9828 \pm 0.0012$
\\
yeast
& $0.6670 \pm 0.0012$ & $\mathbf{0.6679 \pm 0.0012}$
& $\mathbf{0.5653 \pm 0.0012}$ & $0.5650 \pm 0.0012$
\\
flags
& $\mathbf{0.7222 \pm 0.0041}$ & $0.7192 \pm 0.0043$
& $\mathbf{0.6056 \pm 0.0058}$ & $0.6028 \pm 0.0052$
\\
scene
& $0.7860 \pm 0.0020$ & $\mathbf{0.7913 \pm 0.0014}$
& $\mathbf{0.7620 \pm 0.0020}$ & $0.7563 \pm 0.0017$
\\
emotions
& $\mathbf{1.5912 \pm 0.0198}$ & $1.5855 \pm 0.0175$
& $\mathbf{0.5775 \pm 0.0037}$ & $0.5774 \pm 0.0036$
\\ \hline
\end{tabularx}

\begin{tabularx}{1\columnwidth}{*5{>{\centering\arraybackslash}X}} \hline
\multicolumn{1}{c}{Dataset}
 & \multicolumn{2}{c}{Hamming loss $\downarrow$} & \multicolumn{2}{c}{Rank loss $\downarrow$} \\ \hline
 & CSRPE & CSRPE-ext & CSRPE & CSRPE-ext \\ \hline
Corel5k
& $0.0108 \pm 0.0001$ & $\mathbf{0.0106 \pm 0.0000}$
& $490.1698 \pm 1.1959$ & $\mathbf{485.73 \pm 0.88}$
\\
bibtex
& $\mathbf{0.0134 \pm 0.0000}$ & $\mathbf{0.0134 \pm 0.0000}$
& $104.9449 \pm 0.3814$ & $\mathbf{102.7833 \pm 0.3211}$
\\
CAL500
& $0.1651 \pm 0.0005$ & $\mathbf{0.1638 \pm 0.0004}$
& $1304.6118 \pm 4.5735$ & $\mathbf{1303.3491 \pm 4.1772}$
\\
enron
& $0.0500 \pm 0.0003$ & $\mathbf{0.0491 \pm 0.0002}$
& $34.3199 \pm 0.1815$ & $\mathbf{33.4665 \pm 0.2060}$
\\
medical
& $0.0100 \pm 0.0002$ & $\mathbf{0.0098 \pm 0.0001}$
& $\mathbf{5.3300 \pm 0.0676}$ & $5.4147 \pm 0.0808$
\\
genbase
& $0.0014 \pm 0.0001$ & $\mathbf{0.0013 \pm 0.0001}$
& $\mathbf{0.3526 \pm 0.0304}$ & $0.3603 \pm 0.0315$
\\
yeast
& $\mathbf{0.1891 \pm 0.0006}$ & $\mathbf{0.1891 \pm 0.0006}$
& $8.4511 \pm 0.0298$ & $\mathbf{8.4481 \pm 0.0255}$
\\
flags
& $0.2585 \pm 0.0038$ & $\mathbf{0.2580 \pm 0.0034}$
& $\mathbf{3.0101 \pm 0.0470}$ & $3.0500 \pm 0.0496$
\\
scene
& $\mathbf{0.0821 \pm 0.0007}$ & $0.0822 \pm 0.0008$
& $0.6793 \pm 0.0083$ & $\mathbf{0.6453 \pm 0.0061}$
\\
emotions
& $\mathbf{0.1994 \pm 0.0022}$ & $0.1996 \pm 0.0023$
& $\mathbf{0.5911 \pm 0.0014}$ & $0.5921 \pm 0.0016$
\\ \hline
\end{tabularx}

\end{table*}

%The experimental setup is the same as for Experiment \ref{convergence}.
The results, which contain the mean and standard error (ste) of the
criteria, are listed in Table \ref{tab:perfect}.
The results demonstrate that CSRPE-ext is slightly better performing, but
the improvement is at best marginal and insignificant.
Even in the CAL500 dataset, where all the label vectors in training and
testing sets are different, there is only a small performance difference between
CSRPE and CSRPE-ext.
The result verifies that our choice of $\mathcal{\tilde{Y}}$ as all the distinct label
vectors in the training set is sufficiently good.

\subsection{Comparison with Other Algorithms}
%\myparagraph{Comparison with Other MLC Algorithms.}
In this experiment, we compare the performance of various MLC and CSMLC
algorithms.
For the MLC competitors, we include different codes applied within ML-ECC
framework.
The competing codes include the \textit{Hamming on repetition code} (HAMR),
\textit{repetition code (REP)}, and \textit{RAKEL repetition code (RREP)}
\cite{CF2013}.
REP and RREP are equivalent to BR~\cite{DBLP:reference/dmkdh/TsoumakasKV10} and
RAKEL~\cite{DBLP:conf/ecml/TsoumakasV07}, respectively.
In addition, CC~\cite{read2011classifier} is added to serve as a baseline competitor together with REP
and RREP.
For CSMLC algorithms, we compete with PCC~\cite{DBLP:conf/icml/DembczynskiCH10}
and CFT~\cite{DBLP:conf/icml/LiL14}.

The results are shown in Table \ref{tab:comp_mlc} and \ref{tab:ttest_results}.
The results show that CSMLC algorithms generally outperform traditional
MLC algorithms.
This justifies that it is important to take cost information into account.
Among the CSMLC algorithms, CSRPE is superior over all other competitors with
respect to F1 and Accuracy score.
For Rank loss, PCC performs slightly better, but CSRPE still performs
competitively with PCC and CFT.
%Each algorithm wins on some datasets.
%Such result justifies CSRPE as a top choice in terms of performance for
%solving CSMLC problems.
Such result justifies CSRPE as a top performing CSMLC algorithm.

\begin{table*}[t]
\centering
\caption{Experiment results (mean $\pm$ ste) on 
different criteria
%F1 score $\uparrow$, Accuracy score $\uparrow$, and Rank loss $\downarrow$
(best in bold)}
\label{tab:comp_mlc}
%\vspace{0.5em}
\tiny
\begin{tabular}{lccccccc}
%\begin{tabularx}{.95\textwidth}{@{} *8{>{\centering\arraybackslash}X}@{}}
\hline
\multicolumn{8}{c}{{\bf F1 score} $\uparrow$} \\ \hline
Dataset & REP (BR) & RREP (RAKEL) & HAMR & CC & PCC & CFT & CSRPE \\ \hline
Corel5k & $.0683 \pm .0011$ & $.1028 \pm .0010$ & $.0608 \pm .0008$ & $.0661 \pm .0009$ & $.1759 \pm .0008$ & $.1708 \pm .0017$ & $\mathbf{.2455 \pm .0012}$  \\
CAL500 & $.3388 \pm .0014$ & $.3527 \pm .0011$ & $.3152 \pm .0012$ & $.3354 \pm .0024$ & $.3540 \pm .0018$ & $.3815 \pm .0016$ & $\mathbf{.4083 \pm .0017}$  \\
bibtex & $.3636 \pm .0009$ & $.3761 \pm .0010$ & $.3658 \pm .0008$ & $.3569 \pm .0009$ & $.3736 \pm .0011$ & $.3957 \pm .0015$ & $\mathbf{.4663 \pm .0008}$  \\
enron & $.5441 \pm .0026$ & $.5336 \pm .0025$ & $.5459 \pm .0023$ & $.5492 \pm .0022$ & $.5508 \pm .0014$ & $.5530 \pm .0013$ & $\mathbf{.5911 \pm .0014}$  \\
medical & $.7883 \pm .0028$ & $.7757 \pm .0034$ & $.7877 \pm .0031$ & $.7924 \pm .0035$ & $.8131 \pm .0023$ & $.7970 \pm .0031$ & $\mathbf{.8203 \pm .0023}$  \\
genbase & $.9897 \pm .0012$ & $.9893 \pm .0014$ & $.9896 \pm .0012$ & $.9896 \pm .0012$ & $\mathbf{.9911 \pm .0007}$ & $.9845 \pm .0009$ & $.9878 \pm .0008$  \\
yeast & $.6119 \pm .0014$ & $.6130 \pm .0011$ & $.6171 \pm .0015$ & $.5968 \pm .0018$ & $.6013 \pm .0013$ & $.6111 \pm .0024$ & $\mathbf{.6670 \pm .0012}$  \\
flags & $.6954 \pm .0045$ & $.6965 \pm .0044$ & $.7005 \pm .0044$ & $.6973 \pm .0048$ & $.7075 \pm .0038$ & $.6725 \pm .0055$ & $\mathbf{.7222 \pm .0041}$  \\
scene & $.5895 \pm .0026$ & $.5926 \pm .0019$ & $.6365 \pm .0021$ & $.6547 \pm .0019$ & $.7306 \pm .0016$ & $.6592 \pm .0027$ & $\mathbf{.7860 \pm .0020}$  \\
emotions & $.5968 \pm .0038$ & $.5773 \pm .0047$ & $.6100 \pm .0035$ & $.6205 \pm .0035$ & $.6384 \pm .0033$ & $.6015 \pm .0043$ & $\mathbf{.6655 \pm .0035}$  \\
\hline
\multicolumn{8}{c}{{\bf Accuracy score} $\uparrow$} \\ \hline
Dataset & REP (BR) & RREP (RAKEL) & HAMR & CC & PCC & CFT & CSRPE \\ \hline
Corel5k & $.0471 \pm .0007$ & $.0696 \pm .0006$ & $.0408 \pm .0009$ & $.0471 \pm .0007$ & $.1135 \pm .0005$ & $.0790 \pm .0019$ & $\mathbf{.1664 \pm .0009}$  \\
CAL500 & $.2097 \pm .0010$ & $.2179 \pm .0008$ & $.1925 \pm .0007$ & $.2085 \pm .0018$ & $.2209 \pm .0012$ & $.2425 \pm .0015$ & $\mathbf{.2645 \pm .0013}$  \\
bibtex & $.3063 \pm .0009$ & $.3103 \pm .0009$ & $.3094 \pm .0008$ & $.3031 \pm .0010$ & $.2940 \pm .0010$ & $.3235 \pm .0011$ & $\mathbf{.3926 \pm .0011}$  \\
enron & $.4303 \pm .0023$ & $.4215 \pm .0022$ & $.4344 \pm .0024$ & $.4437 \pm .0021$ & $.4259 \pm .0013$ & $.4363 \pm .0018$ & $\mathbf{.4772 \pm .0016}$  \\
medical & $.7559 \pm .0034$ & $.7431 \pm .0033$ & $.7604 \pm .0033$ & $.7643 \pm .0035$ & $.7716 \pm .0025$ & $.7570 \pm .0031$ & $\mathbf{.7939 \pm .0024}$  \\
genbase & $.9859 \pm .0014$ & $.9852 \pm .0015$ & $.9856 \pm .0014$ & $.9858 \pm .0014$ & $\mathbf{.9873 \pm .0009}$ & $.9792 \pm .0012$ & $.9835 \pm .0010$  \\
yeast & $.5047 \pm .0014$ & $.5065 \pm .0012$ & $.5120 \pm .0015$ & $.4954 \pm .0021$ & $.4872 \pm .0017$ & $.5027 \pm .0019$ & $\mathbf{.5653 \pm .0012}$  \\
flags & $.5849 \pm .0047$ & $.5860 \pm .0046$ & $.5913 \pm .0051$ & $.5908 \pm .0057$ & $.5974 \pm .0041$ & $.5616 \pm .0059$ & $\mathbf{.6056 \pm .0058}$  \\
scene & $.5791 \pm .0025$ & $.5816 \pm .0020$ & $.6258 \pm .0017$ & $.6457 \pm .0018$ & $.6821 \pm .0019$ & $.6467 \pm .0029$ & $\mathbf{.7620 \pm .0020}$  \\
emotions & $.5179 \pm .0037$ & $.4959 \pm .0045$ & $.5320 \pm .0034$ & $.5417 \pm .0035$ & $.5433 \pm .0035$ & $.5216 \pm .0036$ & $\mathbf{.5775 \pm .0037}$  \\
\hline
\multicolumn{8}{c}{{\bf Hamming loss} $\downarrow$} \\ \hline
Dataset & REP (BR) & RREP (RAKEL) & HAMR & CC & PCC & CFT & CSRPE \\ \hline
Corel5k & $.0095 \pm .0000$ & $.0097 \pm .0000$ & $\mathbf{.0094 \pm .0000}$ & $.0095 \pm .0000$ & $.0095 \pm .0000$ & $.0100 \pm .0000$ & $.0108 \pm .0001$  \\
CAL500 & $.1522 \pm .0010$ & $\mathbf{.1416 \pm .0003}$ & $.1490 \pm .0005$ & $.1493 \pm .0006$ & $.1493 \pm .0006$ & $.1422 \pm .0005$ & $.1651 \pm .0005$  \\
bibtex & $\mathbf{.0124 \pm .0000}$ & $.0130 \pm .0000$ & $.0124 \pm .0000$ & $.0125 \pm .0000$ & $.0125 \pm .0000$ & $.0136 \pm .0000 $ & $.0134 \pm .0000$  \\
enron & $.0489 \pm .0002$ & $.0499 \pm .0002$ & $.0485 \pm .0002$ & $.0506 \pm .0002$ & $.0506 \pm .0002$ & $\mathbf{.0477 \pm .0002}$ & $.0500 \pm .0003$  \\
medical & $.0104 \pm .0002$ & $.0107 \pm .0001$ & $.0102 \pm .0002$ & $.0103 \pm .0002$ & $.0105 \pm .0001$ & $.0111 \pm .0002$ & $\mathbf{.0100 \pm .0002}$  \\
genbase & $.0012 \pm .0001$ & $.0011 \pm .0001$ & $\mathbf{.0011 \pm .0001}$ & $.0011 \pm .0001$ & $.0012 \pm .0001$ & $.0016 \pm .0001$ & $.0014 \pm .0001$  \\
yeast & $.1941 \pm .0007$ & $.1933 \pm .0006$ & $.1932 \pm .0007$ & $.2040 \pm .0009$ & $.2043 \pm .0009$ & $.2080 \pm .0010$ & $\mathbf{.1891 \pm .0006}$  \\
flags & $.2591 \pm .0037$ & $.2591 \pm .0027$ & $.2599 \pm .0037$ & $.2611 \pm .0037$ & $.2636 \pm .0038$ & $.2899 \pm .0040$ & $\mathbf{.2585 \pm .0038}$  \\
scene & $.0914 \pm .0004$ & $.0970 \pm .0005$ & $.0848 \pm .0005$ & $.0919 \pm .0007$ & $.0924 \pm .0007$ & $.1031 \pm .0009$ & $\mathbf{.0821 \pm .0007}$  \\
emotions & $.1966 \pm .0021$ & $.2110 \pm .0022$ & $\mathbf{.1953 \pm .0019}$ & $.1959 \pm .0018$ & $.1958 \pm .0020$ & $.2207 \pm .0020$ & $.1994 \pm .0022$  \\
\hline
\multicolumn{8}{c}{{\bf Rank loss} $\downarrow$} \\ \hline
Dataset & REP (BR) & RREP (RAKEL) & HAMR & CC & PCC & CFT & CSRPE \\ \hline
Corel5k & $618.1 \pm .6695$ & $597.2 \pm .6664$ & $623.5 \pm .6474$ & $636.0 \pm .5374$ & $421.2 \pm .6626$ & $\mathbf{300.7 \pm .7848}$ & $490.2 \pm 1.1959$  \\
CAL500 & $1500. \pm 5.023$ & $1477. \pm 4.835$ & $1537. \pm 4.488$ & $1520. \pm 6.155$ & $1179. \pm 4.498$ & $\mathbf{1122. \pm 4.470}$ & $1305. \pm 4.574$  \\
bibtex & $132.6 \pm .2981$ & $124.1 \pm .2511$ & $131.5 \pm .2819$ & $136.8 \pm .2886$ & $\mathbf{69.10 \pm .2454}$ & $112.06 \pm .2811$ & $104.9 \pm .3814$  \\
enron & $43.39 \pm .2919$ & $44.06 \pm .2810$ & $43.40 \pm .2540$ & $43.56 \pm .3000$ & $27.94 \pm .1681$ & $\mathbf{27.20 \pm .1365}$ & $34.32 \pm .1815$  \\
medical & $5.454 \pm .1184$ & $5.733 \pm .1088$ & $5.601 \pm .1232$ & $5.469 \pm .0997$ & $\mathbf{3.058 \pm .0603}$ & $4.117 \pm .0741$ & $5.330 \pm .0676$  \\
genbase & $.2461 \pm .0281$ & $.2422 \pm .0273$ & $.2525 \pm .0257$ & $.2423 \pm .0308$ & $\mathbf{.1976 \pm .0178}$ & $.4686 \pm .0310$ & $.3863 \pm .0341$  \\
yeast & $9.609 \pm .0358$ & $9.565 \pm .0290$ & $9.443 \pm .0312$ & $10.324 \pm .0448$ & $9.378 \pm .0365$ & $9.473 \pm .0363$ & $\mathbf{8.451 \pm .0298}$  \\
flags & $3.123 \pm .0434$ & $3.139 \pm .0383$ & $3.078 \pm .0352$ & $3.120 \pm .0450$ & $3.012 \pm .0490$ & $3.363 \pm .0504$ & $\mathbf{3.010 \pm .0470}$  \\
scene & $1.136 \pm .0066$ & $1.149 \pm .0055$ & $1.031 \pm .0046$ & $1.098 \pm .0080$ & $0.726 \pm .0060$ & $0.892 \pm .0069$ & $\mathbf{0.679\pm .0083}$  \\
emotions & $1.789 \pm .0182$ & $1.906 \pm .0220$ & $1.764 \pm .0165$ & $1.741 \pm .0207$ & $\mathbf{1.563 \pm .0176}$ & $1.834 \pm .0281$ & $1.591 \pm .0198$  \\
\hline
\end{tabular}
%\end{tabularx}
%\vspace{-1.5em}
\end{table*}

%\begin{table}[t]
%\center
%\caption{CSRPE versus others based on $t$-test at 95\% confident level}
%\label{tab:ttest_results}
%%\scriptsize
%\tiny
%\begin{tabular}{ccccc} \hline
%criteria (win/tie/loss) & F1 & Rank.  & Acc.   & total  \\ \hline
%REP  & 9/1/0 & 7/2/1  & 9/0/1 & 27/7/6   \\ \hline
%RREP & 9/1/0 & 9/0/1  & 9/1/0 & 31/5/4   \\ \hline
%HAMR & 9/1/0 & 7/2/1  & 8/2/0 & 26/9/5   \\ \hline
%CC   & 9/1/0 & 7/2/1  & 8/2/0 & 30/6/4   \\ \hline
%CFT  & 9/1/0 & 6/1/3  & 9/1/0 & 30/4/6   \\ \hline
%PCC  & 9/0/1 & 2/2/6  & 8/1/1 & 22/7/11  \\ \hline
%\end{tabular}
%\end{table}

\begin{table}[h]
\centering
\caption{CSRPE versus others based on $t$-test at 95\% confident level (win/tie/loss)}
\label{tab:ttest_results}
\scriptsize
\begin{tabularx}{1\columnwidth}{@{} *7{>{\centering\arraybackslash}X}@{}} \hline
criteria & REP(BR) & RREP(RAKEL) & HAMR   & CC     & CFT    & PCC    \\ \hline
f1       & 9/1/0   & 9/1/0       & 9/1/0  & 9/1/0  & 9/1/0  & 9/0/1  \\ \hline
acc.     & 9/0/1   & 9/1/0       & 8/2/0  & 8/2/0  & 9/1/0  & 8/1/1  \\ \hline
hamming  & 2/4/4   & 4/3/3       & 2/4/4  & 6/1/3  & 6/1/3  & 3/4/3  \\ \hline
rank.    & 7/2/1   & 9/0/1       & 7/2/1  & 7/2/1  & 6/1/3  & 2/2/6  \\ \hline
total    & 27/7/6  & 31/5/4      & 26/9/5 & 30/6/4  & 30/4/6 & 22/7/11  \\ \hline
\end{tabularx}
\end{table}

\subsection{Comparison with MLAL Algorithms}
%\myparagraph{Comparison with MLAL Algorithms}
In this experiment, we evaluate the performance of CSRPE under the CSMLAL
setting.
We compare it with several state-of-the-art MLAL algorithms, which includes
\textit{adaptive active learning} (adaptive)~\cite{li2013active},
\textit{maximal loss reduction with maximal confidence}
(MMC)~\cite{yang2009effective}, and random sampling as a baseline algorithm.
Their implementations were obtained from \textsc{libact}~\cite{libact}.
We do not include a comparison with \textit{binary
minimization}~\cite{brinker2006active} since MMC and adaptive are
reported to outperform it.

\begin{figure}
%\vspace{-1.5em}
\centering

\subfloat[CAL500]{\includegraphics[width=.30\textwidth]{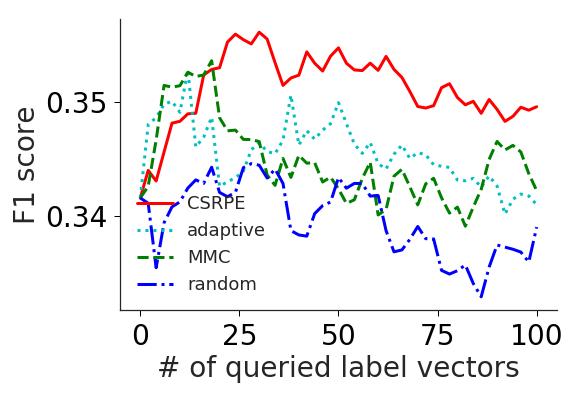}}
\subfloat[enron]{\includegraphics[width=.30\textwidth]{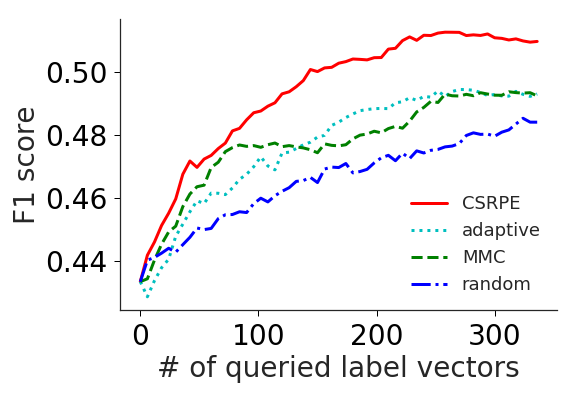}}
\subfloat[medical]{\includegraphics[width=.30\textwidth]{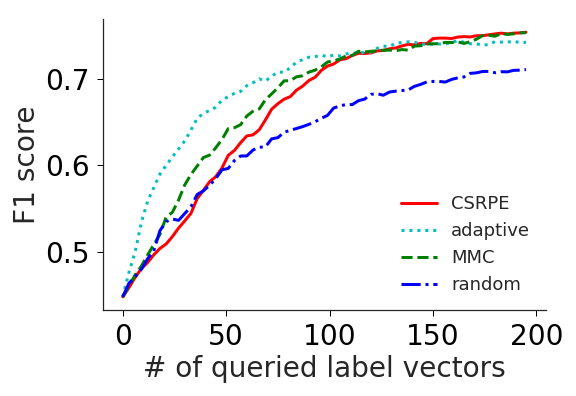}}

%\vspace{-1.em}
\subfloat[yeast]{\includegraphics[width=.30\textwidth]{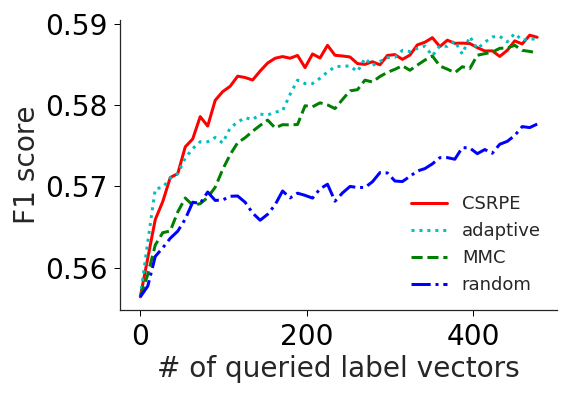}}
\subfloat[scene]{\includegraphics[width=.30\textwidth]{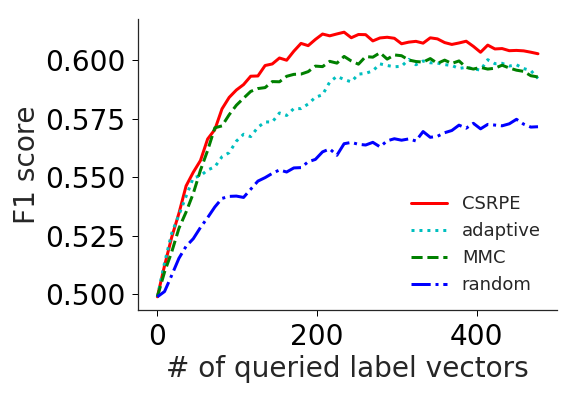}}
\subfloat[emotions]{\includegraphics[width=.30\textwidth]{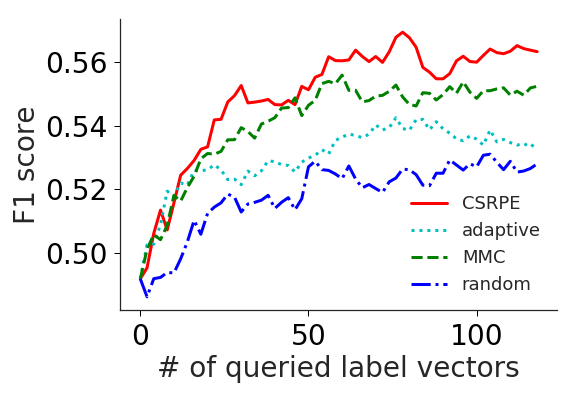}}

\caption{CSMLAL results with F1 score $\uparrow$}
\label{fig:al_f1}
%\vspace{-1.0em}
\end{figure}
\begin{figure}[!t]
%\vspace{-1.5em}
\centering

\subfloat[CAL500]{\includegraphics[width=.30\textwidth]{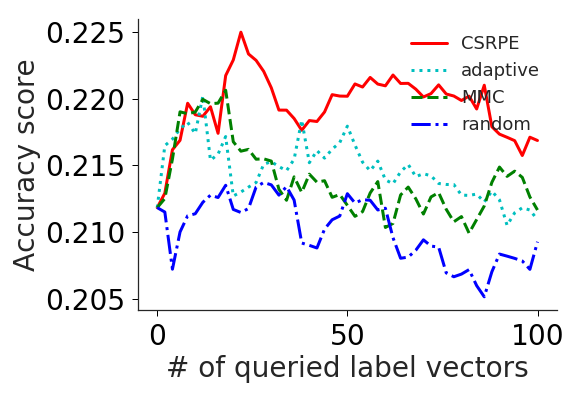}}
\subfloat[enron]{\includegraphics[width=.30\textwidth]{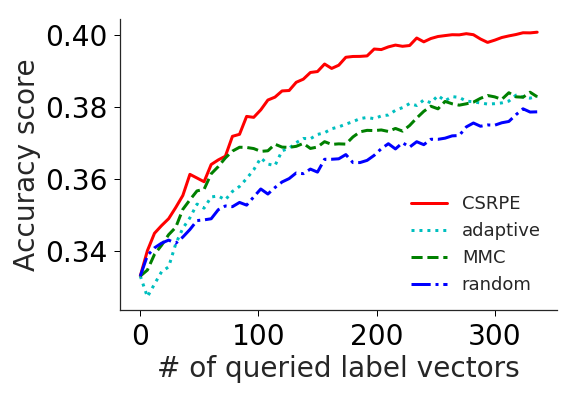}}
\subfloat[medical]{\includegraphics[width=.30\textwidth]{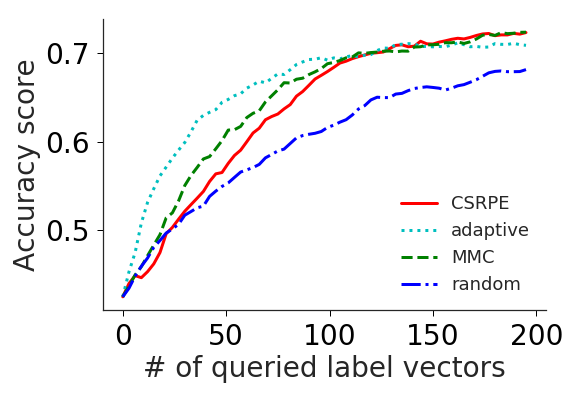}}

%\vspace{-1.em}
\subfloat[yeast]{\includegraphics[width=.30\textwidth]{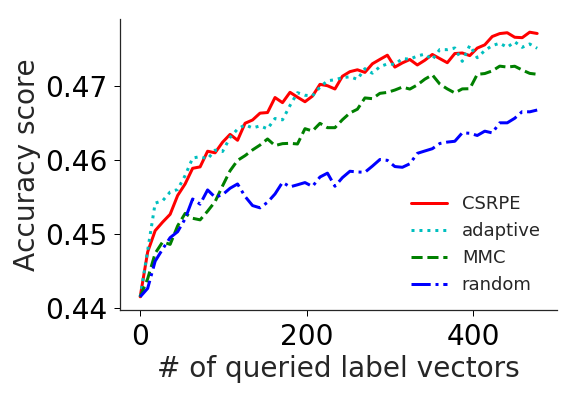}}
\subfloat[scene]{\includegraphics[width=.30\textwidth]{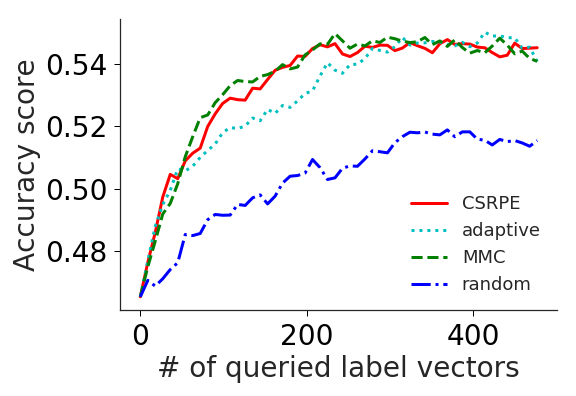}}
\subfloat[emotions]{\includegraphics[width=.30\textwidth]{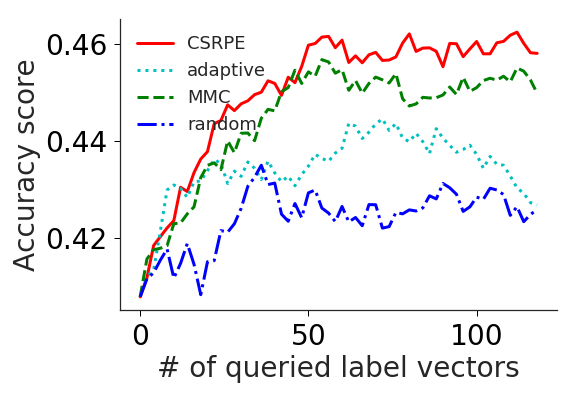}}

\caption{CSMLAL results with Accuracy score $\uparrow$}
\label{fig:al_acc}
%\vspace{-1.0em}
\end{figure}

\begin{figure}
\centering

\subfloat[CAL500]{\includegraphics[width=.30\textwidth]{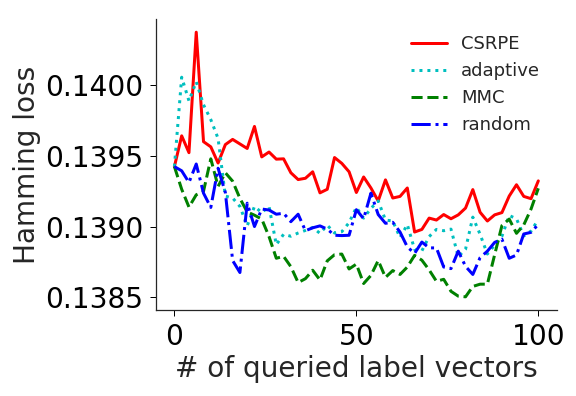}}
\subfloat[enron]{\includegraphics[width=.30\textwidth]{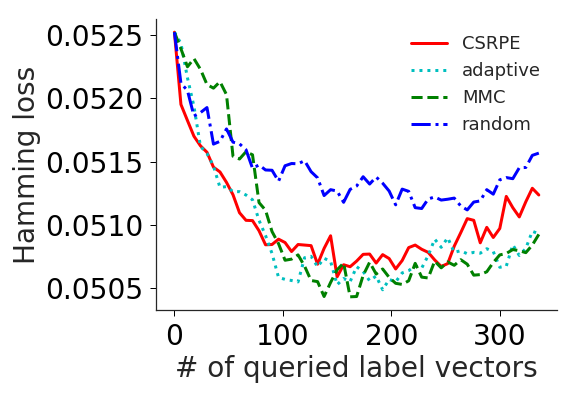}}
\subfloat[medical]{\includegraphics[width=.30\textwidth]{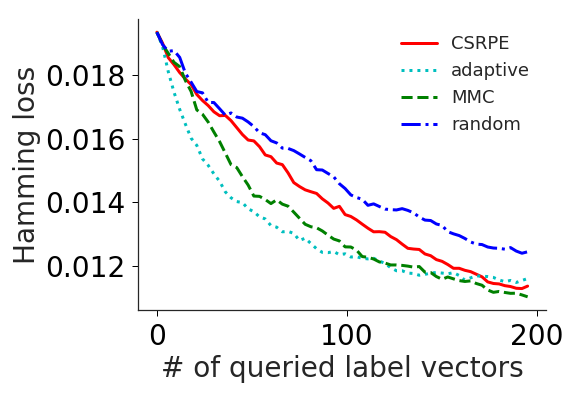}}

\subfloat[yeast]{\includegraphics[width=.30\textwidth]{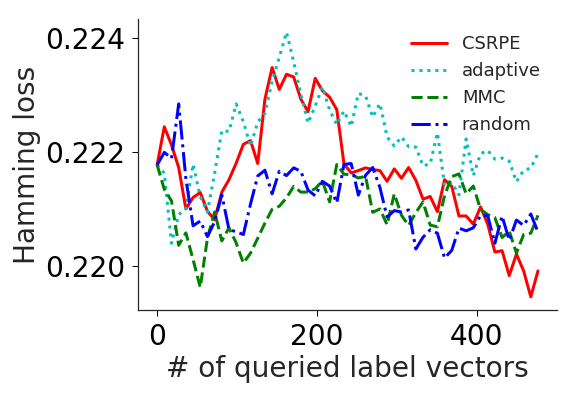}}
\subfloat[scene]{\includegraphics[width=.30\textwidth]{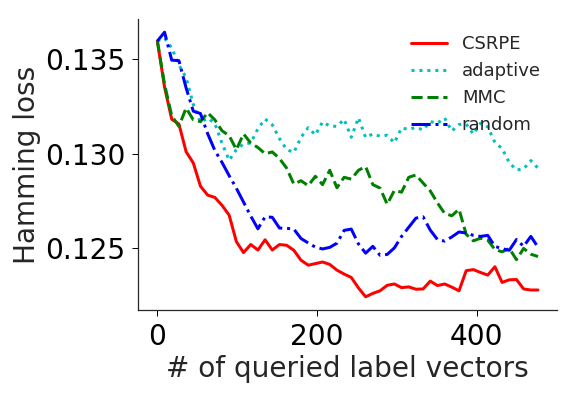}}
\subfloat[emotions]{\includegraphics[width=.30\textwidth]{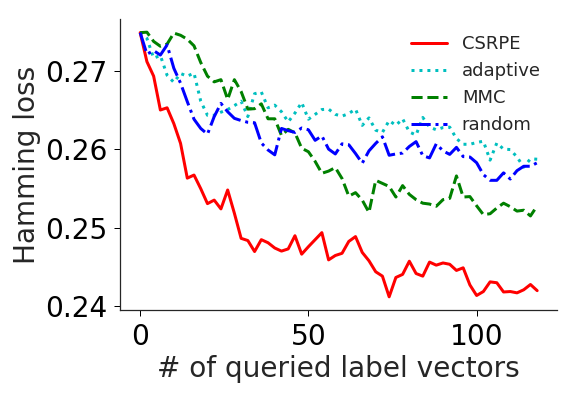}}

\caption{CSMLAL results with Hamming loss $\downarrow$}
\label{fig:al_hamming}
%%\vspace{-1.5em}
\end{figure}

\begin{figure}
%\vspace{-1.5em}
\centering

\subfloat[CAL500]{\includegraphics[width=.30\textwidth]{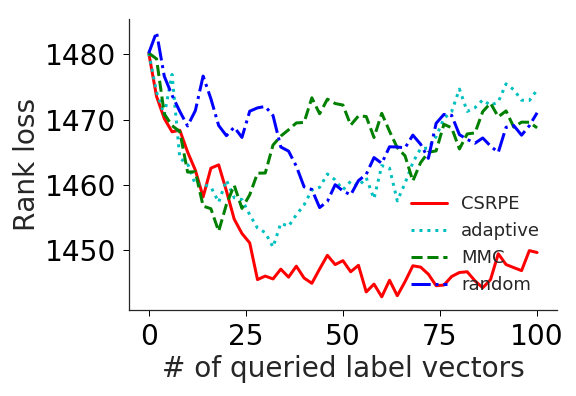}}
\subfloat[enron]{\includegraphics[width=.30\textwidth]{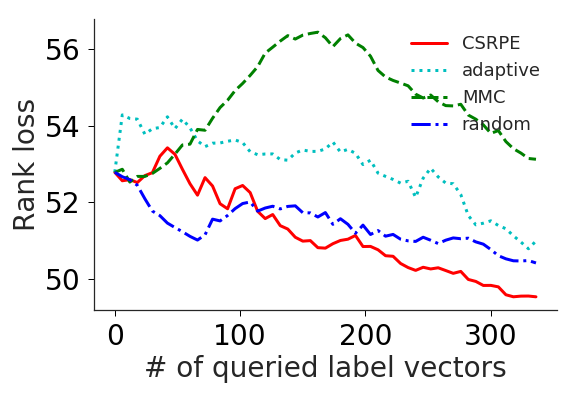}}
\subfloat[medical]{\includegraphics[width=.30\textwidth]{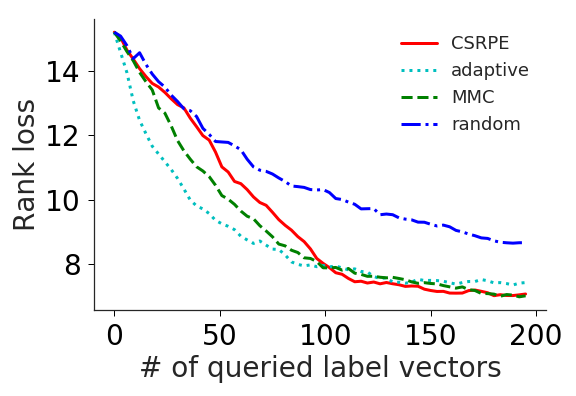}}

%\vspace{-1.em}
\subfloat[yeast]{\includegraphics[width=.30\textwidth]{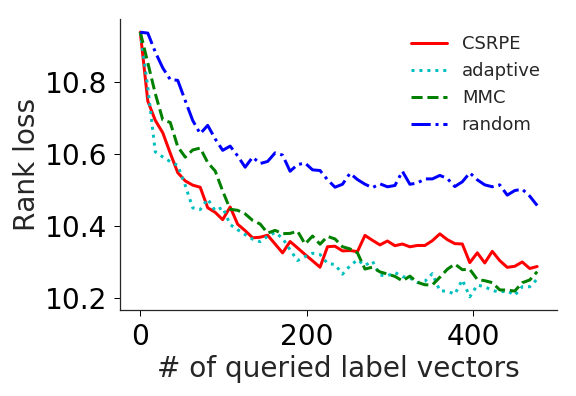}}
\subfloat[scene]{\includegraphics[width=.30\textwidth]{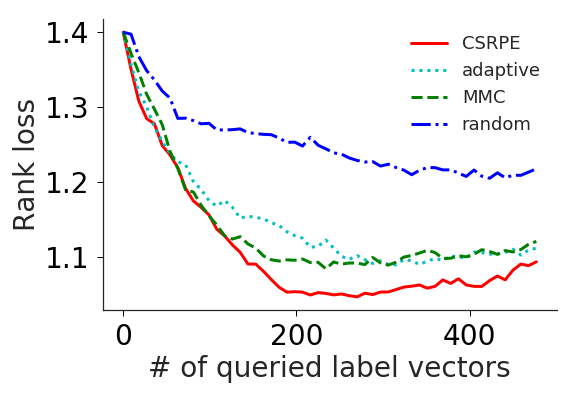}}
\subfloat[emotions]{\includegraphics[width=.30\textwidth]{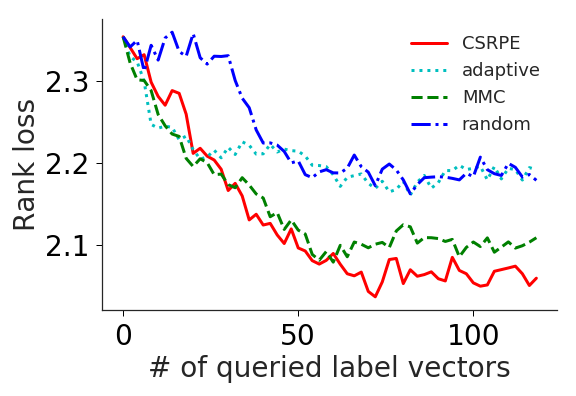}}

\caption{CSMLAL results with Rank loss $\downarrow$}
\label{fig:al_rank}
%\vspace{-1.0em}
\end{figure}

%Figure \ref{fig:al_f1_acc} shows the
Figures \ref{fig:al_f1}, \ref{fig:al_acc}, \ref{fig:al_hamming}, and \ref{fig:al_rank}
show the performance with respect to the number of instances queried.
For F1 score and Rank loss, CSRPE performs better than other
strategies on four out of six datasets.
%On the yeast dataset, CSRPE performs competitively with other strategies and
%the best results for the medical dataset are obtained by the adaptive strategy.
%These results indicate that CSRPE is able to consider the cost information during
%the querying process, thus enabling it to outperform other competitors on most of
%the datasets across different evaluation criteria.
These results indicate that CSRPE is able to consider the cost information,
thus enabling it to outperform other competitors on most of the datasets
across different evaluation criteria.

\section{Conclusion} \label{conclusion}
In this paper, we propose a novel approach for cost-sensitive
multi-label classification (CSMLC), called \textit{cost-sensitive reference
pair encoding}~(CSRPE).
CSRPE is derived from the one-versus-one algorithm and can embed the cost
information into the encoded vectors.
Exploiting the redundancy of the encoded vectors, we use random sampling to
resolve the training challenge of building so many classifiers.
We also design a nearest-neighbor-based decoding procedure and use the
relevant set to efficiently make cost-sensitive predictions.
Extensive experimental results demonstrate that CSRPE achieves stable
convergence respect to the code length and outperforms not only other
encoding methods but also state-of-the-art CSMLC algorithms across
different cost functions.
In addition, we extend CSRPE to a novel multi-label active learning algorithm
by designing a cost-sensitive uncertainty measure.
Extensive empirical studies show that the proposed active learning algorithm
performs better than existing active learning algorithms.
The results suggest that CSRPE is a promising cost-sensitive encoding method
for CSMLC for either supervised or active learning.

%\subsection*{Acknowledgments.}
\section*{Acknowledgments}
We thank the anonymous reviewers and the members of NTU CLLab for valuable
suggestions.
This material is based
upon work supported by the Air Force Office of Scientific
Research, Asian Office of Aerospace Research and Development
(AOARD) under award number FA2386-15-1-4012,
and by the Ministry of Science and Technology of Taiwan
under MOST 103-2221-E-002-149-MY3 and 106-2119-M-007-027.

\clearpage

\bibliographystyle{splncs}
\bibliography{csrpe}

\end{document}